\documentclass[10pt,journal,compsoc]{IEEEtran}
\ifCLASSOPTIONcompsoc
  \usepackage[nocompress]{cite}
\else
  \usepackage{cite}
\fi
\usepackage{amsmath,amsfonts}
\usepackage{bbm}
\usepackage{multirow}
\usepackage{booktabs}
\usepackage{url}
\usepackage[noend]{algorithmic}
\usepackage{algorithm}
\usepackage{graphicx}
\usepackage{subfigure}
\usepackage{color}
\usepackage[table]{xcolor}
\usepackage{pifont}

\ifCLASSINFOpdf
\else
\fi
\begin{document}
%
\title{Instance Consistency Regularization for Semi-Supervised 3D Instance Segmentation}
%
%
%
%

\author{Yizheng Wu\textsuperscript{*}, 
        Zhiyu Pan\textsuperscript{*},
        Kewei Wang,
        Xingyi Li,
        Jiahao Cui,
        Liwen Xiao,
        Guosheng Lin,
        Zhiguo Cao
\IEEEcompsocitemizethanks{\IEEEcompsocthanksitem Yizheng Wu, Kewei Wang, and Xingyi Li are with the Key Laboratory
of Image Processing and Intelligent Control, Ministry of Education, School of Artificial Intelligence and Automation, Huazhong University of Science and Technology, Wuhan 430074, China, and also with S-Lab, Nanyang Technological University, Singapore 639798, Singapore. Email: yzwu21@hust.edu.cn, wangkewei@hust.edu.cn, xingyi\_li@hust.edu.cn
\IEEEcompsocthanksitem Zhiyu Pan, Jiahao Cui, Liwen Xiao, and Zhiguo Cao are with the Key Laboratory
of Image Processing and Intelligent Control, Ministry of Education, School of Artificial Intelligence and Automation, Huazhong University of Science and Technology, Wuhan 430074, China. Email: zhiyupan@hust.edu.cn, smartcjh@hust.edu.cn, liwenxiao@hust.edu.cn, zgcao@hust.edu.cn
\IEEEcompsocthanksitem Guosheng Lin is with Nanyang Technological University, Singapore 639798, Singapore. Email: gslin@ntu.edu.sg
\IEEEcompsocthanksitem Yizheng Wu and Zhiyu Pan contributed equally.

}
}

%
%

\markboth{}
{}
\IEEEtitleabstractindextext{%
\begin{abstract}
Large-scale datasets with point-wise semantic and instance labels are crucial to 3D instance segmentation but also expensive. To leverage unlabeled data, previous semi-supervised 3D instance segmentation approaches have explored self-training frameworks, which rely on high-quality pseudo labels for consistency regularization. They intuitively utilize both instance and semantic pseudo labels in a joint learning manner. However, semantic pseudo labels contain numerous noise derived from the imbalanced category distribution and natural confusion of similar but distinct categories, which leads to severe collapses in self-training. Motivated by the observation that 3D instances are non-overlapping and spatially separable, we ask whether we can solely rely on instance consistency regularization for improved semi-supervised segmentation. To this end, we propose a novel self-training network InsTeacher3D to explore and exploit pure instance knowledge from unlabeled data. We first build a parallel base 3D instance segmentation model DKNet, which distinguishes each instance from the others via discriminative instance kernels without reliance on semantic segmentation. Based on DKNet, we further design a novel instance consistency regularization framework to generate and leverage high-quality instance pseudo labels. Experimental results on multiple large-scale datasets show that the InsTeacher3D significantly outperforms prior state-of-the-art semi-supervised approaches. Code is available: \url{https://github.com/W1zheng/InsTeacher3D}. 
\end{abstract}

}

\maketitle

\IEEEdisplaynontitleabstractindextext
\begin{figure*}[!h]
\centering
\subfigure[Comparison of instance and semantic segmentation.]{\includegraphics[width=0.43\linewidth]{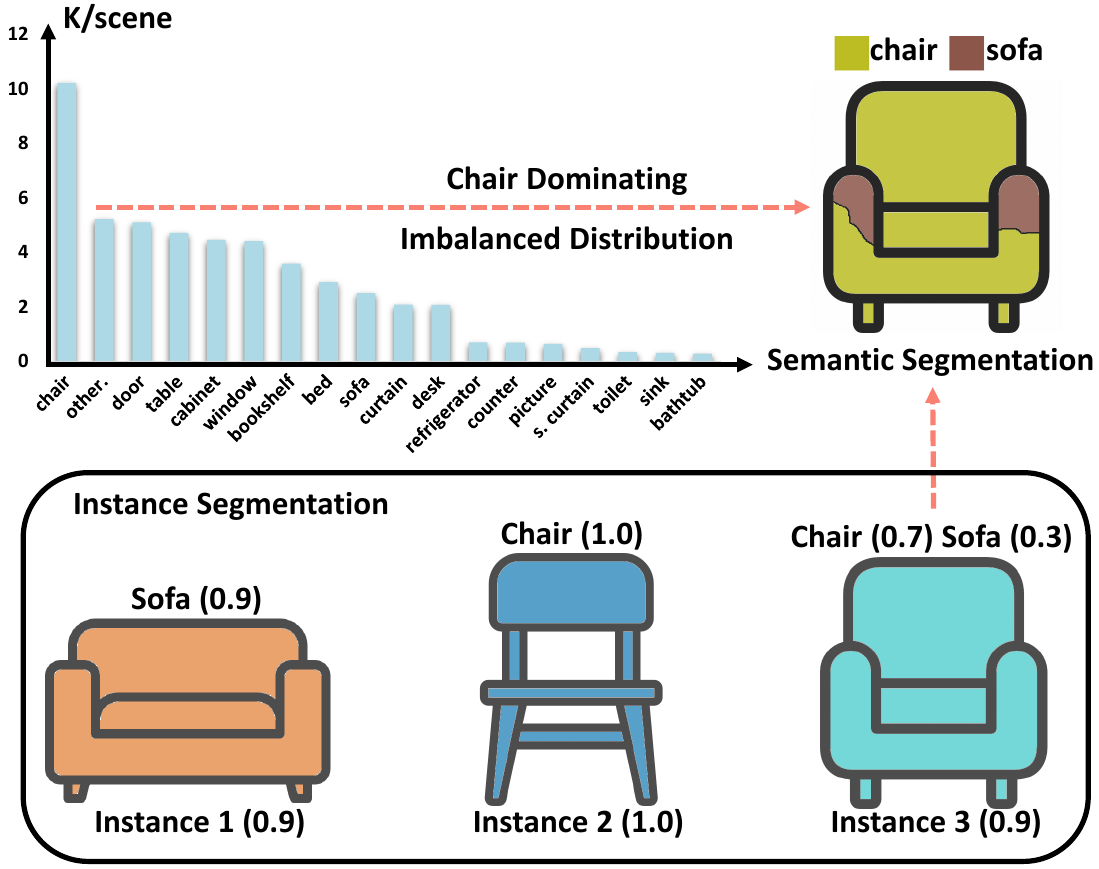}}
\hfill
\rule{1pt}{6cm}
\hfill
\subfigure[Fuzzy semantic pseudo labels.]{\includegraphics[width=0.27\linewidth]{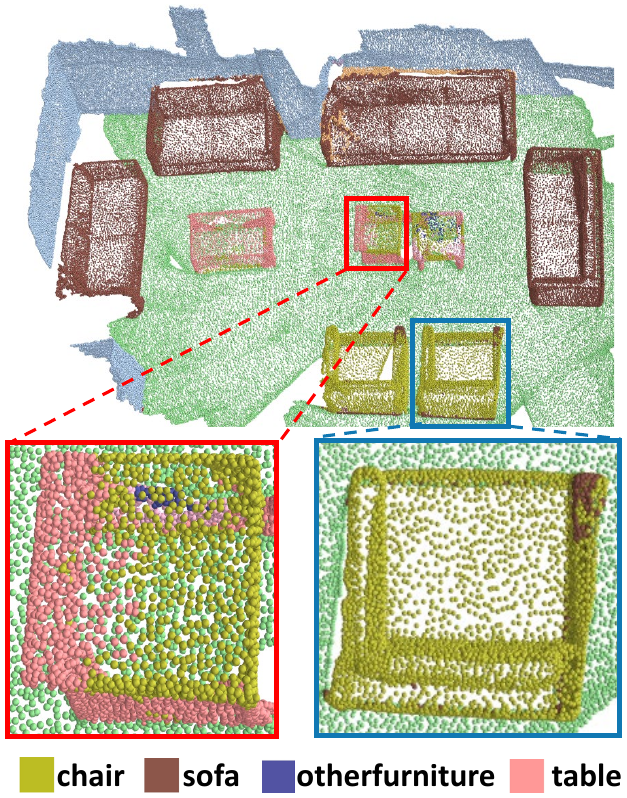}}
\subfigure[Sharp instance pseudo labels.]{\includegraphics[width=0.27\linewidth]{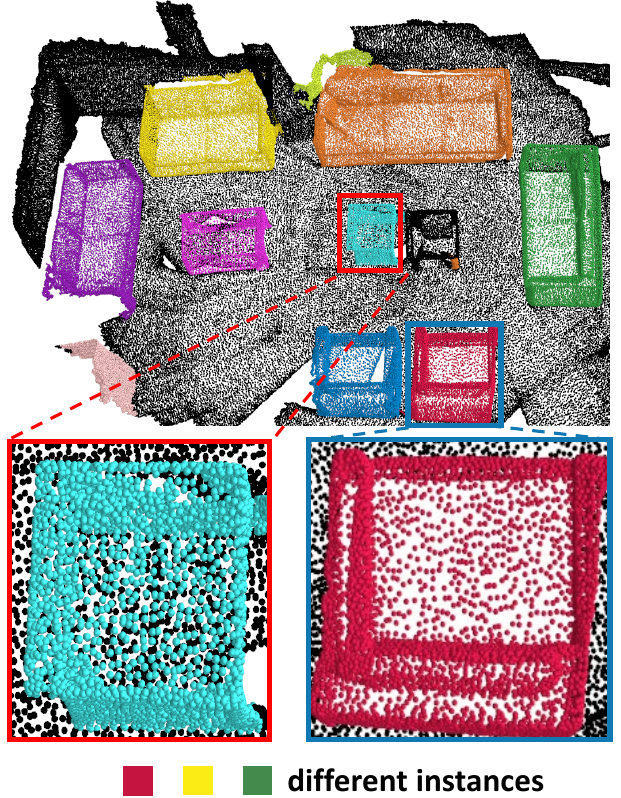}}
\vspace{-0.3cm}
\caption{\textbf{The comparison between semantic and instance pseudo labels.} (a) An armchair combines features of both a sofa and a chair. This confusion, along with the imbalance in semantic categories, results in poor semantic pseudo labels. In contrast, the non-overlapping and spatially separable nature of 3D instances leads to accurate instance pseudo labels. We present the confidence scores within parentheses. (b) The semantic pseudo labels may shatter the instances into multiple semantic parts, which causes ambiguity during the self-training procedure.  (c) Instance pseudo labels are sharp enough and keep instances as cohesive units, which benefits learning from unlabeled data.}
\label{fig:sem-ins}
\vspace{-0.5cm}
\end{figure*}

%
\IEEEpeerreviewmaketitle

\IEEEraisesectionheading{\section{Introduction}\label{sec:introduction}}

%
%
%
%

\IEEEPARstart{3}{D} instance segmentation is critical for 3D scene understanding, and its success highly relies on datasets with complete point-level annotations. However, the annotation process is time-consuming and costly: One worker takes around $22.3$ minutes to annotate dozens of instances with thousands of points in a single scene; For dataset validity, over 500 annotation workers contribute to ScanNetV2~\cite{scannet}, in which each scene needs 2.3 workers on average.
The high cost of annotations is a significant barrier to scaling up 3D instance segmentation datasets, creating a bottleneck for further progress. Therefore, 
boosting 3D instance segmentation 
with the help of unlabeled data is in demand.

Unsupervised and semi-supervised approaches have been explored to address this challenge. Unsupervised works~\cite{pointcontrast,csc} pre-train the model on unlabeled data through contrastive learning and then finetune it on labeled data. However, these unsupervised approaches have limited success in leveraging unlabeled data due to the loose constraints of contrastive loss. Hence, some semi-supervised approaches utilize the self-training paradigm to generate pseudo labels on unlabeled data and learn via consistency regularization. For effective self-training, it is significant to guarantee the quality of pseudo labels. TWIST~\cite{twist} applies a self-training framework combined with a serial segmentation model, where both instance and semantic pseudo labels are leveraged and instance segmentation relies on semantic prediction. Nevertheless, as depicted in Fig.~\ref{fig:sem-ins}(a), the imbalanced long-tail distribution of semantic categories and natural confusion between similar but distinct semantic categories, such as chairs and sofas, render the semantic pseudo labels unreliable. As illustrated in Fig.~\ref{fig:sem-ins}(b), the above factors contribute to one instance being wrongly separated into several semantic parts, which causes a subsequent impact on self-training. In contrast, as 3D instances are non-overlapping and spatially separable in point clouds, we observe that instance pseudo labels remain sharp and cohesive, as shown in Fig~\ref{fig:sem-ins}(c). 
These factors prompt us to question \textit{whether we can solely rely on instance consistency regularization for improved self-training}. 

We find it non-trivial to explore and exploit pure instance knowledge from unlabeled data. Specifically, there are two fundamental challenges to be addressed: 1) how to mitigate the impact of noisy semantic predictions and 2) how to effectively enhance and leverage instance pseudo labels. On the one hand, current state-of-the-art segmentation approaches either rely on heuristic clustering~\cite{pointgroup,hais,softgroup} or transformer-based paradigms~\cite{mask3d,lai2023mask,spformer} for instance segmentation. However, cluster-based methods introduce noises due to their serial reliance on semantic predictions, leading to self-training collapse. Transformer-based approaches with parallel instance query encoding techniques struggle with learning from limited labeled data, hindering their performance under data-efficient settings. On the other hand, enhancing instance pseudo labels for robust consistency regularization is crucial. While TWIST~\cite{twist} attempts a two-way inter-label enhancement using both instance and semantic masks, it fails when semantic pseudo labels are omitted for pure instance consistency regularization.

To remedy these issues, we propose InsTeacher3D, a novel self-training network to leverage instance consistency regularization to improve semi-supervised 3D instance segmentation. At the base model level, we build a kernel-based 3D instance segmentation model DKNet, performing instance segmentation parallel to the noisy semantic segmentation branch. This architecture enables InsTeacher3D to mitigate the impact of noisy semantic predictions. Then, for stable self-training, InsTeacher3D discards semantic pseudo labels and enhances the instance pseudo labels by injecting task-specific priors to facilitate instance consistency regularization. With high-quality instance pseudo labels, InsTeacher3D employs a student model to learn potential instance knowledge from unlabeled data, and the weight of the student model is used to update the teacher model via exponential moving average (EMA)~\cite{mt} for robust pseudo-labeling and consistency regularization. The above procedure allows InsTeacher3D to focus on learning discriminative instance knowledge, without being influenced by semantic ambiguity.

We conduct experiments to evaluate the performance of our InsTeacher3D on three popular datasets: ScanNetV2~\cite{scannet}, S3DIS~\cite{s3dis}, and STPLS3D~\cite{chen2022stpls3d}.
Our results show that InsTeacher3D significantly outperforms previous state-of-the-art self-training instance segmentation approaches. Moreover, it ranks first on the online hidden test set of the data-efficient ScanNetV2 benchmark.\footnote[1]{3D Semantic Instance with Limited Reconstructions Benchmark at: \url{https://kaldir.vc.in.tum.de/scannet_benchmark/data_efficient/index?percent_scenes=20percent&metric=ap}}
In conclusion, our study emphasizes that exploiting unlabeled data via instance consistency regularization is better for semi-supervised 3D instance segmentation.

Our contributions are four-fold:
\begin{itemize}
    \item We identify that, for the semi-supervised 3D instance segmentation, the origin of noise in pseudo labels is semantic ambiguity.
    \item We design a parallel 3D instance segmentation approach DKNet, the pivot of which is building discriminative instance kernels to mitigate the impact of noisy semantic predictions. 
    \item We propose InsTeacher3D as a self-training network designed to facilitate instance consistency regularization. InsTeacher3D generates and leverages high-quality instance pseudo labels to effectively explore potential instance information from unlabeled data. 
    \item We conduct a series of experiments and ablation studies to prove the effectiveness of our approach and evaluate both the instance and semantic pseudo labels. InsTeacher3D thoroughly explores the more reliable instance consistency and outperforms the previous approach by a large margin.
\end{itemize}

The preliminary version of this work was presented at the European Conference on Computer Vision (ECCV) 2022, entitled “3D Instances as 1D Kernels”. The conference version showcases the strong instance discrimination capability of proposed DKNet, demonstrating significant potential for semi-supervised learning. This journal submission contains substantial extensions to semi-supervised 3D instance segmentation based on DKNet.  We also include thorough analysis, algorithmic enhancements, as well as more extensive experiments and discussions in this version.

\section{Related Work}\label{sec:related-work}
\subsection{3D Instance Segmentation}
As a significant task in 3D scene understanding, 3D instance segmentation assigns instance and semantic labels for every point. 
The proposal-based approaches~\cite{gspn,bonet,gicn,3dmpa} first detect object proposals and then separate the foreground and background points within proposals. In contrast, proposal-free approaches~\cite{sgpn,sstnet} group the points into instances based on embedding similarity. 
The recent works~\cite{pointgroup,hais,softgroup,rpgn} have developed heuristic clustering algorithms that serially rely on semantic categories and latent instance hints to group points, which can be seen as an extension of semantic segmentation involving instance clues.
The transformer-based approaches~\cite{mask3d,spformer,lai2023mask} follow the general framework of the attention-based architecture~\cite{attention} and focus on the design of the query encoder to encode and refine instances queries. 
Nevertheless, initialization and encoding of instance queries rely on fitting large-scale datasets, which makes it not suitable for data-efficient learning. 
On the contrary, kernel-based approaches~\cite{dyco3d,pointinst3d,dknet,isbnet} generate or assign dynamic kernels~\cite{dfn} for instances depending on exploring scenes.
DyCo3D~\cite{dyco3d} built kernels based on coarse grouping proposals; PointInst3D~\cite{pointinst3d} apply farthest point sampling (FPS) to sample points as candidate kernels; ISBNet~\cite{isbnet} introduce instance-aware FPS and further aggregation to obtain kernels. 
We propose DKNet~\cite{dknet} with a localize-then-aggregate paradigm to search local maximums in centroid heatmap as candidates and further aggregate neighbor information of candidates to represent 3D instances as kernels. 
Therefore, DKNet could exploit scene information and instance knowledge to thoroughly explore the data similarity, which enables reconstructing accurate instance masks even with limited labeled data. 
In this work, we employ DKNet as the base segmentation model in InsTeacher3D, which leverages instance localization, representation, and reconstruction consistency for effective semi-supervised learning.

\subsection{Data-Efficient 3D Scene Understanding}
Data-efficient scene understanding concerns learning with limited scenes or annotations to reduce the data collection and annotation cost. In 3D point cloud, these problems can be broadly divided into two settings: Limited Annotation(LA) and Limited Reconstruction(LR)~\cite{csc}, which correspond to weakly supervised and semi-supervised learning tasks respectively. 
Some approaches first explore unsupervised pre-training on unlabeled data and finetuning under a data-efficient setting.
PointContrast~\cite{pointcontrast} introduces point-wise contrastive loss to enhance features. 
CSC~\cite{csc} integrates spatial contexts into pre-training features by introducing contrastive scene contexts. However, these approaches have limited success in leveraging unlabeled data due to the loose constraints of contrastive loss. Further, there are numerous works exploring weakly supervised and semi-supervised approaches in diverse data-efficient 3D perception tasks: object detection~\cite{3dioumatch,sess,Liu_2023_CVPR}, semantic segmentation~\cite{xu2020weakly,otoc,scribble,xu2023hierarchical}, instance segmentation~\cite{box2mask,twist} \emph{et al}.

In this paper, we focus on semi-supervised 3D instance segmentation under the limited reconstruction setting, where only a limited proportion of scenes are fully labeled with others left unlabeled. 
TWIST~\cite{twist} first explores this task and builds a self-training network to leverage unlabeled data. WS3D~\cite{liu2022weakly} proposes unsupervised region-level energy-based and contrastive loss to achieve boundary awareness and instance discrimination. Both approaches build upon PointGroup~\cite{pointgroup} where instance segmentation serially relies on semantic prediction. The semantic categories and coarse centroid offset vectors are employed as pseudo labels. However, due to the serial segmentation manner, these approaches are unable to fully leverage instance information and are plagued by noisy semantic prediction.
Cooperating with our DKNet, InsTeacher3D abandons semantic pseudo labels and solely focuses on exploiting instance consistency from unlabeled data for effective self-training.

\subsection{Self-Training Semi-Supervised Learning}
The self-training approaches generate high-quality pseudo labels for unlabeled data and utilize consistency regularization to leverage potential information in unlabeled data and boost the semi-supervised performance. 
There is a large volume of published studies~\cite{mt,stac,twist,deng2021unbiased,yang2022st++,zou2019confidence,pan2024pseudo,wang2024semi} describing the role of the self-training technique in semi-supervised learning.
However, these approaches usually suffer from noise in consistency regularization, which can be mitigated by further improving the quality of pseudo labels.
Mean teacher~\cite{mt} ensembles student models at every step via EMA as the teacher model, which generates better pseudo labels and guides the student model via consistency regularization. 
TWIST~\cite{twist} proposes a self-training network with object denoising and two-way label enhancement modules to produce semantic and instance pseudo labels. 
In this work, we first identify the origin of noise in pseudo labels and propose a novel self-training strategy relying solely on instance pseudo labels to leverage unlabeled data. To facilitate instance consistency, we develop InsTeacher3D following the general mean teacher framework and further introduce a dynamic mask generation module to produce high-quality instance pseudo labels. 

\section{Preliminary: Consistency Regularization for Semi-Supervised 3D Segmentation} \label{sec:mt}
Given labeled and unlabeled point clouds $P^l$ and $P^u$ respectively, semi-supervised 3D instance segmentation aims to utilize $P^u$ to boost the performance. Throughout this paper, all the superscripts $l$ and $u$ denote the labeled and unlabeled portions, respectively. In this work, we leverage the unlabeled data via consistency regularization, which supposes that different models should produce the same prediction for unlabeled data even with different disturbances. We build our self-training network following the most popular consistency regularization framework, mean teacher~\cite{mt}.

In the mean teacher framework, a teacher $\Phi^t$ produces high-quality pseudo labels, which guide the student $\Phi^s$ by applying consistency regularization upon unlabeled data. We use superscripts $t$ and $s$ to represent the elements of teacher and student models, respectively.
Both $\Phi^t$  and $\Phi^s$  are the same in initial weights and model architectures, but different in data input and weight updating.
Specifically, $\Phi^t$ and $\Phi^s$ process the raw point cloud with weak and strong augmentation, respectively.
Besides, the weights of $\Phi^s$ are updated via stochastic gradient descent while $\Phi^t$ uses the exponential moving average (EMA) weights of the student model. The update of EMA weights of $\Phi^t$ after step $\tau$ can be formulated as:
\begin{equation}
    \label{eq:ema}
    \Phi^t_{\tau + 1} = \alpha \cdot \Phi^t_{\tau} + (1-\alpha) \cdot \Phi^s_{\tau + 1} .
\end{equation}
This self-ensembling procedure can smooth the model noise and improve the performance of $\Phi^t$ to produce better pseudo labels. $\alpha$ is set to $0.999$ as a smoothing factor. 

As stated above, the efficacy of consistency regularization heavily depends on the performance of $\Phi^t$ and the quality of the pseudo labels. If pseudo labels contain numerous noise, the student model will overfit the noisy pseudo label and finally collapse, which is known as confirmation bias~\cite{mt}. Thus, for stable self-training, it is significant to enhance the quality of pseudo labels to alleviate the confirmation bias. In this work, we identify that the origin of confirmation bias in semi-supervised 3D instance segmentation is semantic ambiguity and propose a novel self-training network InsTeacher3D to address it.

\section{InsTeacher3D}

\subsection{Overview of InsTeacher3D}

Semantic ambiguity is a common phenomenon in 3D instance segmentation, especially under semi-supervised settings. As shown in Table~\ref{table:ambiguity}, up to $48.6\%$ instances are with semantic ambiguity, but only a few instances are with instance ambiguity, which demonstrates the instance pseudo labels are more reliable. Thus, the primary objective of InsTeacher3D is to use the power of instance consistency while mitigating the impact of noisy semantic predictions.

\begin{figure}[!t]
\centering
\includegraphics[width=0.9\linewidth]{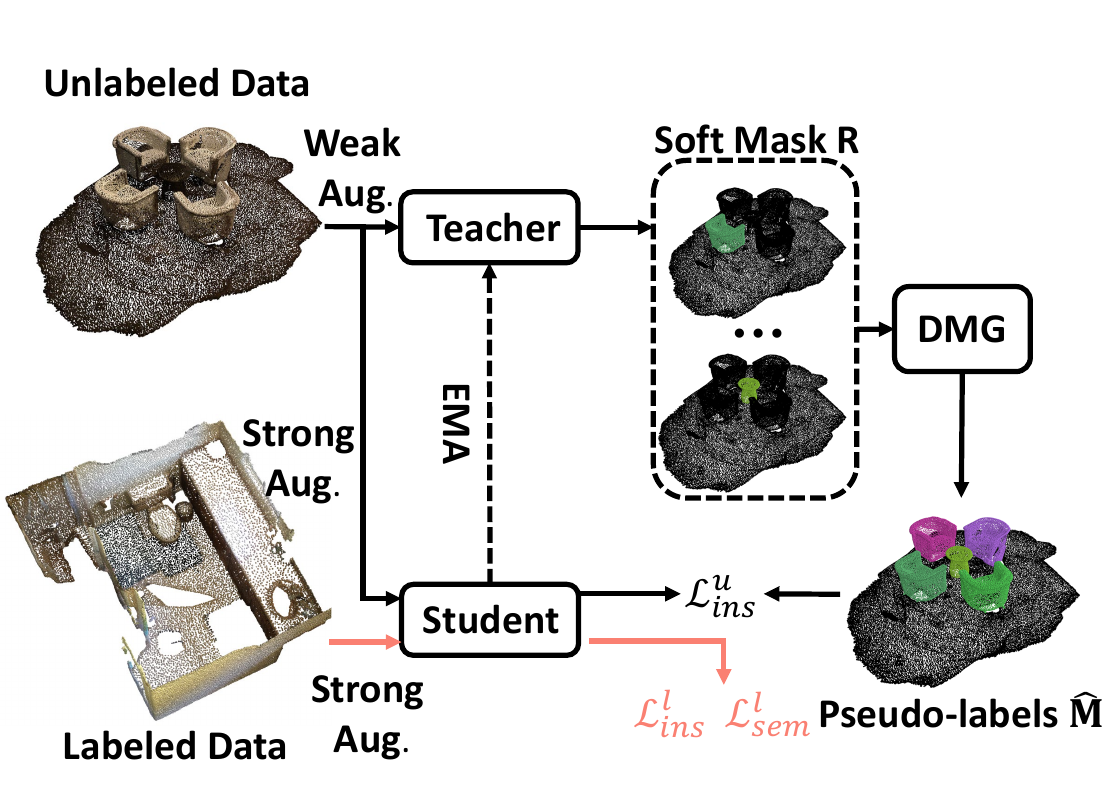}
\vspace{-0.6cm}
\caption{\textbf{The pipeline of InsTeacher3D.} InsTeacher3D is a self-training network in instance consistency regularization framework, where DKNet is the base segmentation model. ``DMG'' denotes the dynamic mask generation module, serving as a key module in instance consistency regularization to generate high-quality instance pseudo labels.}
\label{fig:pipeline-insteacher3d}
\vspace{-0.4cm}
\end{figure}

To this end, we design InsTeacher3D along two dimensions: 1) at the model level, we leverage our proposed parallel instance segmentation model, DKNet to produce instance masks without reliance on semantic prediction; 2) at the semi-supervised learning level, we build an instance consistency regularization framework to generate and leverage high-quality instance pseudo labels to exploit unlabeled data. We present the pipeline of InsTeacher3D in Fig.~\ref{fig:pipeline-insteacher3d}. The base segmentation model (DKNet) and semi-supervised framework (instance consistency regularization) jointly contribute to the success of InsTeacher3D.
In Sec.~\ref{sec:JCR}, we construct a naive transitional network combining DKNet with the intuitive joint consistency regularization (JCR) framework. Notably, we observe that this naive network is still susceptible to noisy semantic pseudo labels stemming from semantic ambiguity, even with strong instance discrimination from DKNet.
Thus, in Sec.~\ref{sec:ICR}, we propose our instance consistency regularization (ICR) framework, which cooperates with a dynamic mask generation module to generate high-quality instance pseudo labels for self-training. As shown in Table~\ref{table:ambiguity_impact}, JCR is hindered by severe semantic ambiguity, while ICR effectively alleviates the impact of semantic ambiguity.

\begin{figure*}[!t]
\centering
\subfigure[JCR (Serial)]{
\includegraphics[width=0.37\linewidth]{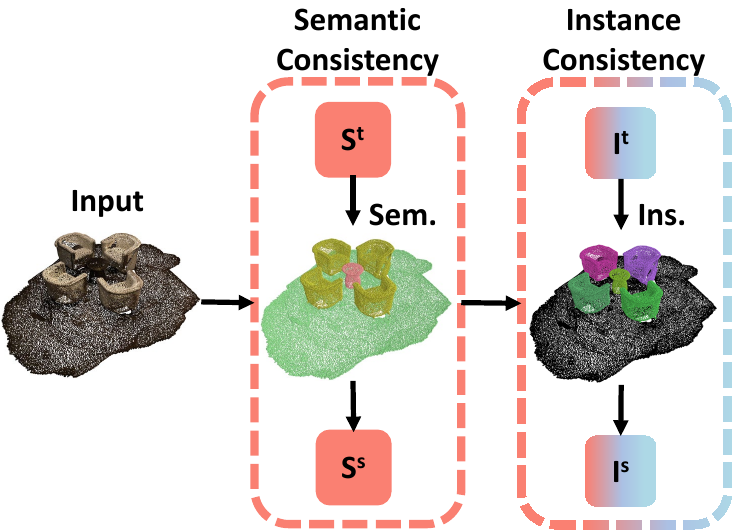}}
\hfill
\rule{1pt}{4.5cm}
\hfill
\subfigure[JCR (Parallel)]{
\includegraphics[width=0.355\linewidth]{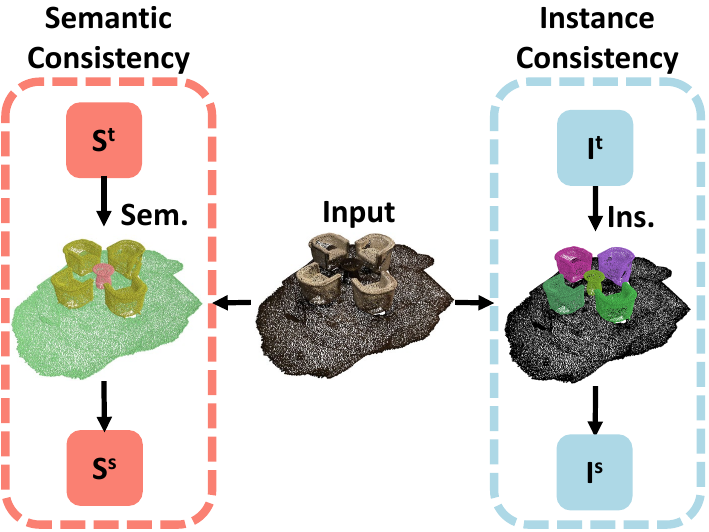}}
\hfill
\rule{1pt}{4.5cm}
\hfill
\subfigure[InsTeacher3D]{\includegraphics[width=0.235\linewidth]{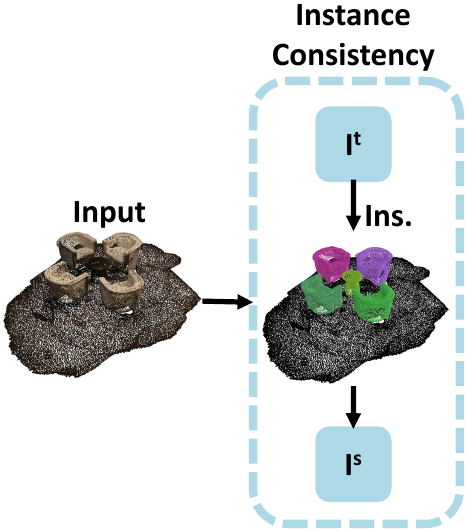}}
\vspace{-0.2cm}
\caption{\textbf{Different architectures for consistency regularization of semi-supervised 3D instance segmentation network.} Here, we focus on the pivotal aspects of consistency regularization on unlabeled data for clarity. The joint consistency regularization (JCR) framework can cooperate with serial and parallel base segmentation models in (a) and (b), respectively. ``I'' and ``S'' represent the instance and semantic segmentation modules. The teacher and student models are denoted by superscripts ``t'' and ``s''. ``Ins.'' and ``Sem.'' are instance and semantic pseudo labels respectively. The red and blue colors represent the knowledge of semantic and instance segmentation. Best viewed in color.}
\label{fig:architecture}
\vspace{-0.4cm}
\end{figure*}

\begin{table}[!t]
\small
\begin{center}
\caption{\textbf{Instance-level evaluation of semantic and instance ambiguity.} We report the instance-level means of point-wise semantic prediction accuracy within each instance and instance segmentation mask IoUs as key metrics. An instance is deemed to exhibit semantic or instance ambiguity if its semantic prediction accuracy or mask IoU falls beneath the $25\%$ threshold. We present the occurrence rates for both types of ambiguity under varying label rates.}
\label{table:ambiguity}
\vspace{-0.4cm}
\begin{tabular*}{0.9\linewidth}{@{}@{\extracolsep{\fill}}c|cccc@{}}
\toprule
\multirow{2}*{Rate}& \multirow{2}*{mAcc}&Semantic&\multirow{2}*{mIoU}&Instance\\
~&~&Ambiguity &~&Ambiguity \\
\midrule
$100\%$&82.1&10.3$\%$&81.2&5.2$\%$\\
$20\%$&76.8&14.9$\%$&78.7&6.5$\%$\\
$10\%$&74.0&17.4$\%$&78.2&6.9$\%$\\
$5\%$&68.5&21.0$\%$&74.4&8.1$\%$\\
$1\%$&42.7&48.6$\%$&53.8&12.8$\%$\\
\bottomrule
\end{tabular*}
\end{center}
\vspace{-0.4cm}
\end{table}
\begin{table}[!t]
\small
\begin{center}
\caption{\textbf{The impact of semantic ambiguity.} We compare JCR against ICR framework with $20\%$ labeled data. The model trained with JCR exhibits a lower instance IoU score and a higher proportion of semantic ambiguity. Meanwhile, the instances with semantic ambiguity tend to be incorrectly segmented, thereby undermining the model performance.}
\label{table:ambiguity_impact}
\vspace{-0.4cm}
\begin{tabular*}{0.9\linewidth}{@{}@{\extracolsep{\fill}}c|ccc@{}}
\toprule
\multirow{2}*{Framework}& \multirow{2}*{mIoU}&Semantic&mIoU\\
~& ~& Ambiguity& (w Ambiguity)\\
\midrule
JCR&73.8&15.6$\%$&58.1\\
ICR&78.7&14.9$\%$&63.1\\
\bottomrule
\end{tabular*}
\end{center}
\vspace{-0.6cm}
\end{table}

\subsection{A Naive Transitional Network}\label{sec:JCR}

\begin{figure}[!t]
\centering
\subfigure[Semantic Segmentation.]{\includegraphics[width=0.48\linewidth]{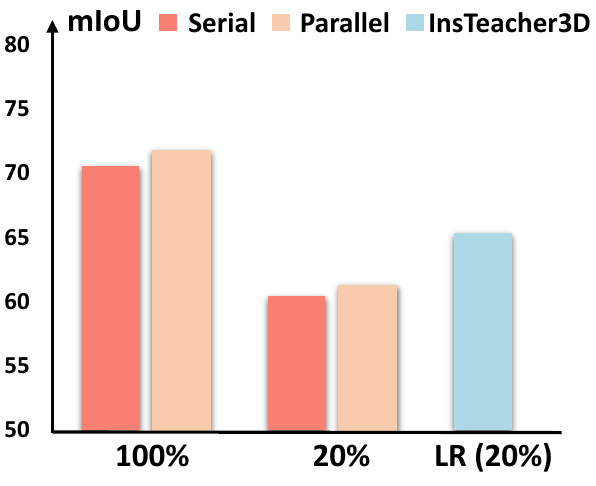}}
\subfigure[Instance Segmentation.]{\includegraphics[width=0.48\linewidth]{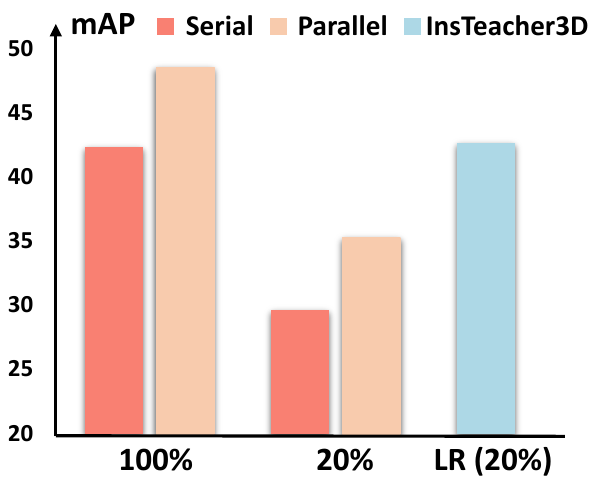}}
\subfigure[Joint regularization.]{\includegraphics[width=0.48\linewidth]{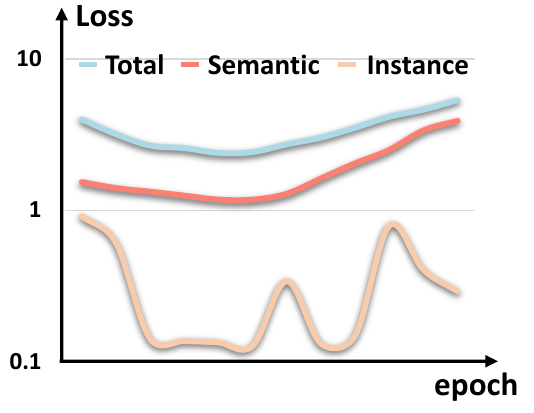}}
\subfigure[Inst. regularization.]{\includegraphics[width=0.48\linewidth]{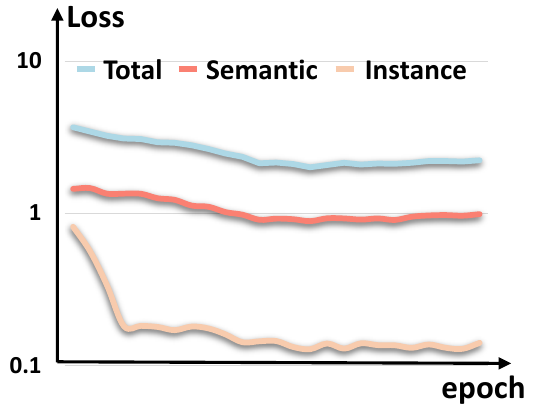}}
\vspace{-0.2cm}
\caption{\textbf{The performance of different instance segmentation architectures and the training curves on S3DIS.} We report the performance of semantic and instance segmentation performance in (a) and (b). HAIS~\cite{hais} and DKNet~\cite{dknet} are adopted as serial and parallel base models, respectively. $100\%$ and $20\%$ denote the two supervised settings, where models are trained with different amounts of labeled data. ``LR ($20\%$)'' is the semi-supervised setting where we train the model with labels from only $20\%$ scenes.  The curves in (c) and (d) represent the loss terms on S3DIS validation sets with joint and instance consistency regularization frameworks, respectively.}
\label{fig:arch-performance}
\vspace{-0.6cm}
\end{figure}

Previous approaches~\cite{twist} adopt joint consistency regularization frameworks, where both semantic and instance pseudo labels are intuitively utilized for regularization. However, as depicted in Fig.~\ref{fig:architecture}(a), if serial segmentation models serve as base models in a joint consistency regularization framework, the dominating noisy semantic prediction leads to unstable self-training. We argue that DKNet explores informative instance representation from scratch rather than relying on semantic prediction, which can be leveraged to alleviate the impact of noisy semantic prediction. The comparison in Fig.~\ref{fig:arch-performance}(a) and (b) proves that DKNet can obtain superior performance of instance segmentation with comparable semantic prediction under two supervised settings ($100\%$ and $20\%$), revealing its superiority in instance discrimination. 

Thus, we conduct a naive transitional network as shown in Fig.~\ref{fig:architecture}(b), which improves the joint consistency regularization framework into a parallel architecture by integrating DKNet as the base segmentation model. However, while a parallel architecture with DKNet leads to remarkable improvement, the joint consistency regularization still suffers from inevitable noisy semantic pseudo labels.
As shown in Fig.~\ref{fig:arch-performance}(c), the noisy semantic pseudo labels result in unstable model learning. Subsequently, the instance segmentation is affected by the model in chaos.
Motivated by the above observation, we design an instance consistency regularization framework, which discards semantic pseudo labels and solely relies on high-quality instance pseudo labels to leverage unlabeled data. 
With this instance consistency regularization framework, InsTeacher3D maintains a stable learning procedure in Fig.~\ref{fig:arch-performance}(d), and effectively boosts the segmentation performance shown in Fig.~\ref{fig:arch-performance}(a) and (b).

\subsection{Instance Consistency Regularization} \label{sec:ICR}
We propose further boosting semi-supervised segmentation through instance consistency regularization, focusing on utilizing only instance pseudo labels to explore potential knowledge from unlabeled data. Our evaluation of instance consistency regularization encompasses two key aspects:
\textit{Instance consistency is sufficient for semi-supervised instance segmentation.} As semantic instances consist of shape, spatial, and category attributions, finding and representing instances facilitate not only instance but also semantic discrimination. Cooperating with semantic supervision on labeled data, the model combines the instance clusters and category names to build complete instance segmentation;
\textit{Pure instance pseudo label is more available.} As every instance can be recognized as an individual class, instance segmentation can get rid of class imbalance in data-efficient learning. Besides, 3D instances are cohesive units and are usually separated by void interspace, which can be exploited for accurate segmentation. 

Based on the efficacy of instance consistency, we introduce InsTeacher3D as a concise self-training network incorporating instance consistency regularization framework, as depicted in Fig.~\ref{fig:pipeline-insteacher3d}. Since this framework relies solely on instance pseudo labels, it is thus crucial to implement an appropriate self-training technique to harness the potential of instance consistency. To achieve this, we designed a dynamic mask generation module and a self-training pipeline specifically for generating and utilizing high-quality instance pseudo labels, respectively.

\begin{figure*}[!t]
\centering
\includegraphics[width=0.9\linewidth]{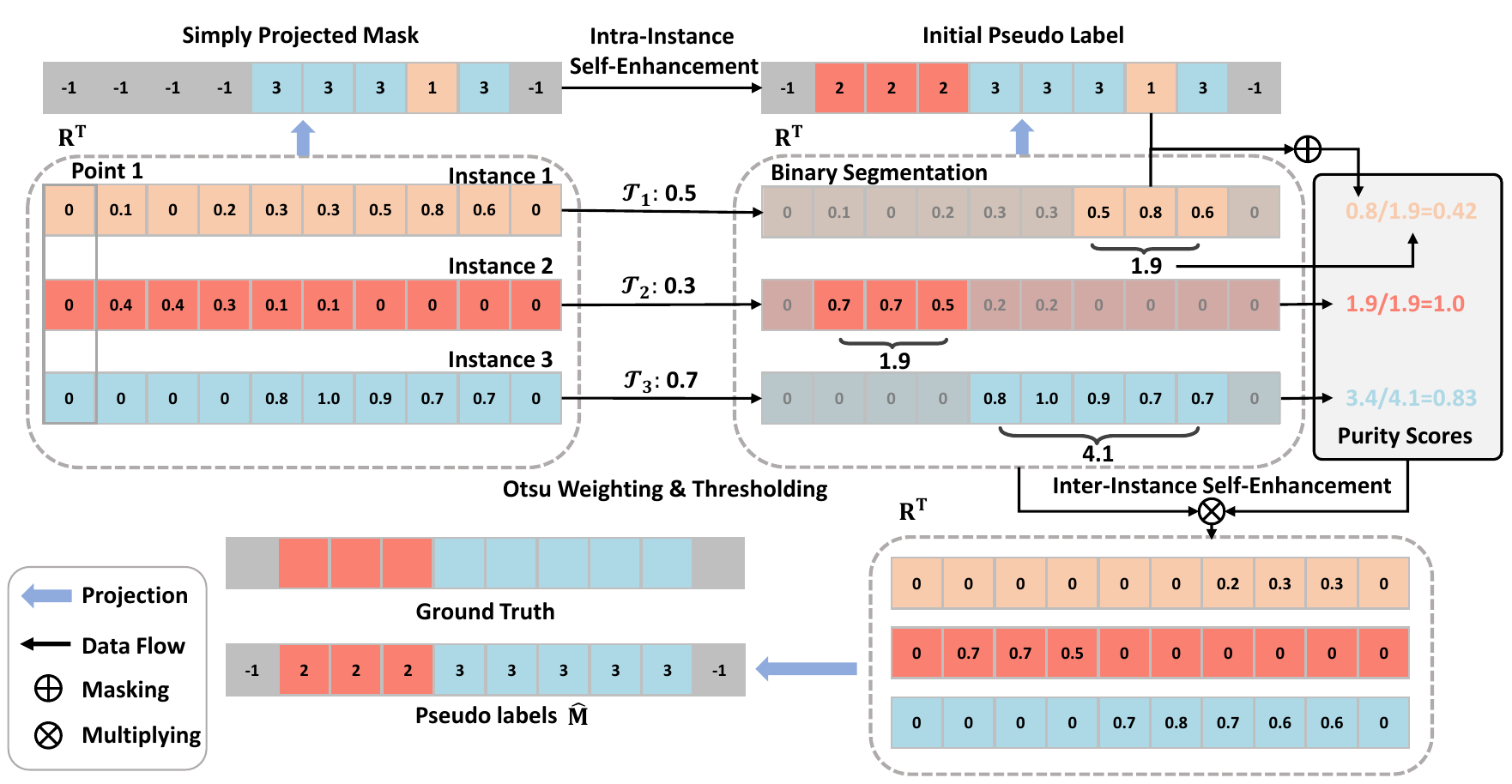}
\vspace{-0.4cm}
\caption{\textbf{A detailed procedure for the instance self-enhancement module.} The simply projected masks contain numerous noises, which will lead to unstable self-training. The proposed instance self-enhancement module effectively improves the quality of pseudo labels $\hat{M}$. }
\label{fig:dmg}
\vspace{-0.4cm}
\end{figure*}

\subsubsection{Dynamic mask generation}\label{sec:dmg}
The teacher model $\Phi^t$, an instance segmentation model implemented as DKNet, produces sparse soft instance masks $R \in \mathbb{R}^{N\times I}$ as initial segmentation results, where $R_{n,i}$ indicates the possibility that point $n$ belongs to instance $i$. However, in these soft masks, one point may respond to multiple kernels, and some instances may show low segmentation confidence and be ignored, leading to noisy segmentation and failing to serve as pseudo labels. 
Thus, we expect to obtain high-quality dense hard instance IDs $\hat{M} = \{\hat{M}_n\in\{-1,1,2,\cdots, I\}\}_{n=1}^N$ as pseudo labels. To this end, we introduce instance self-enhancement to exploit the relationship priors between soft and hard instance masks to preserve weak instances and filter out noisy instances to generate pseudo labels. Superpoint~\cite{superpoint} refinement is also utilized to enhance the completeness of pseudo labels.

\noindent\textbf{Instance self-enhancement.}
Firstly, to ensure one point belongs to one instance, we should project the soft masks into hard point-wise instance IDs. 
Since $R_{n,i}$ indicates the possibility that point $n$ belongs to instance $i$, the most intuitive mask transform is the channel-wise maximum projection, which can be formulated as:
\begin{equation}\label{eq:maximum-projection}
\hat{M}_n = \left\{ 
    \begin{aligned}
        &\mathop{\arg\max}_{i}(R_{n,i}) &  max(R_n)>0.5, foreground\\
        &-1  &  max(R_n)\leq0.5, background\\
    \end{aligned}
\right. ,
\end{equation}
where $\hat{M}_n$ and $R_n$ denote instance ID and instance similarity vector of the $n^{th}$ point, respectively. The instance ID $-1$ denotes the background. In this case, all the instances share the same weight, and all points with low confidence are recognized as backgrounds, which may ignore the weak instances, even for well-learning models.
Therefore, we propose dynamic thresholding and weighting techniques to preserve the weak instances in projection as an intra-instance self-enhancement.
Focusing on single instance, ${R^T}_i$ is the point-wise binary segmentation mask for $i^{th}$ instance.
In ${R^T}_i$, since scores of points belonging to $i^{th}$ instance should clearly overweight other points, a dynamic threshold $\mathcal{T}_i$ can be used for binary segmentation.
Specifically, we implement a maximum between-class variance (Otsu)~\cite{otsu} algorithm to evaluate $\mathcal{T}_i$, which can indicate the intensity of the instance, \emph{i.e}. weak instances should have low thresholds. 
Then, the inverse of the dynamic threshold is utilized as a weight to balance the instance score distribution, which can be formulated as ${R^T}_i = {R^T}_i/2\mathcal{T}$. This processing effectively boosts the scores of weak instances, and we also limit the maximum dynamic thresholds to $0.5$ to avoid suppressing the strong objects.
Here, $R$ is projected according to Eqn.\ref{eq:maximum-projection} to build initial $\hat{M}$.

\begin{algorithm}[!t]  
  \caption{Instance Self-Enhancement} 
  \label{alg: Targets}  
  \begin{algorithmic}[1]  
    \REQUIRE  Instance soft masks $R \in \mathbb{R}^{N\times I}$
    \ENSURE Instance pseudo labels $\hat{M}$\\
    \STATE \textcolor{cyan}{/* Intra-Instance Self-Enhancement*/}
    \FOR{$i \in [1,\ I]$} 
        \STATE $\mathcal{T} \leftarrow \ {\rm Otsu}({R^T}_i)$
        \STATE \textbf{if} $\mathcal{T} < 0.5$ \textbf{then} Update ${R^T}_i$ to ${R^T}_i\ /\ 2\mathcal{T}$
    \ENDFOR
    \STATE $\hat{M} \leftarrow  {\rm Projection}(R)$ according to Eqn.~\ref{eq:maximum-projection}
    \STATE \textcolor{cyan}{/* Inter-Instance Self-Enhancement*/}
    \FOR{$i \in [1,\ I]$}     
        \STATE Calculate $S_{purity}$ according to Eqn.~\ref{eq:noisy_score}
        \STATE Update ${R^T}_i$ to ${R^T}_i \cdot S_{purity}$
    \ENDFOR
    \STATE $\hat{M} \leftarrow  {\rm Projection}(R)$ according to Eqn.~\ref{eq:maximum-projection} only for foreground points
    \STATE return $\hat{M}$
  \end{algorithmic}  
\end{algorithm}

\begin{figure*}[!t]
\centering
\includegraphics[width=0.7\linewidth]{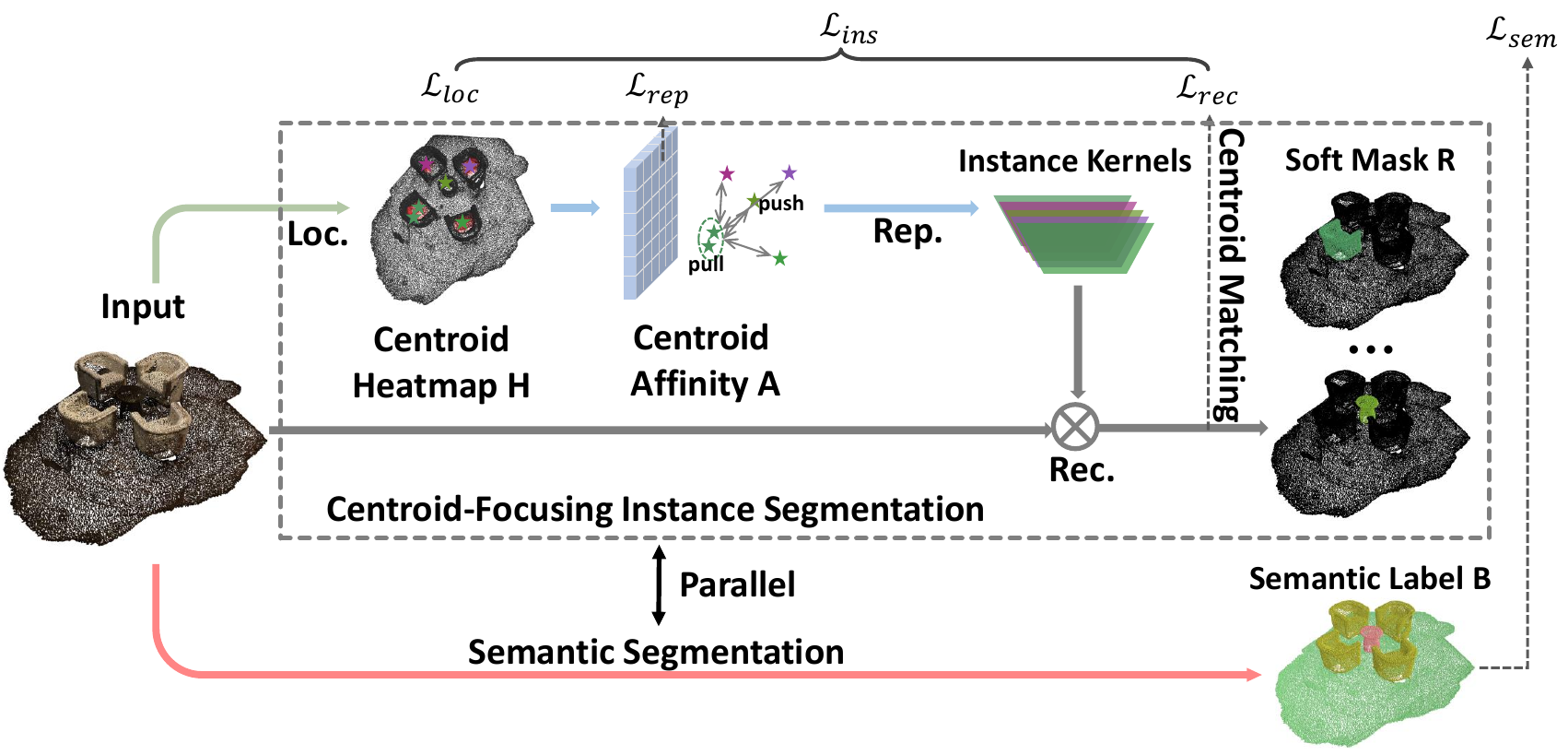}
\vspace{-0.4cm}
\caption{\textbf{The pipeline of DKNet.} DKNet separates the instance segmentation with semantic category assignment as a parallel segmentation framework. ``Loc.'', ``Rep.'', and ``Rec.'' represent instance localization, representation, and reconstruction module. All the above modules focus on instance centroid area to alleviate the potential noise.}
\label{fig:pipeline-dknet}
\vspace{-0.6cm}
\end{figure*}

In addition, though we have tried to build one-to-one correspondence by merging redundant candidates, there are still multiple masks recovered for a single instance. 
To filter out noisy masks, we exploit the interaction between soft and hard instance masks for inter-instance self-enhancement.
For pure instances, the binary segmented soft masks (only concerning one mask) should produce the same result as hard instance IDs (involving all masks).
Thus, we define purity score $S_{purity}$ as:
\begin{equation} \label{eq:noisy_score}
    S_{purity} = \frac{\sum_{n=1}^N {R_{n,i}\cdot \mathbbm{1}(\hat{M}_n=i)}}{\sum_{n=1}^N{R_{n,i}\cdot \mathbbm{1}(R_{n,i}>0.5)}},
\end{equation}
where $\mathbbm{1}(\cdot)$ is an indicator function.
This score reflects how noisy the instance masks are, \emph{i.e}. the scores of pure instances with the same results in both soft and hard masks will be $1$ while the instances disappearing in $\hat{M}$ will get $0$.
Then we can suppress the noisy instances via multiplying $S_{purity}$ with ${R^T}_i$ and re-project the foreground points in $R$ to further enhance $\hat{M}$.

A detailed procedure for the instance self-enhancement module is illustrated in Fig.~\ref{fig:dmg}. In this example, the original soft masks $R$ will produce noisy results after projection. The intra-instance self-enhancement highlights the weak instance (instance 2), and the inter-instance self-enhancement suppresses the duplicated instance (instance 1). The enhanced soft masks are finally projected to the high-quality pseudo labels.

\noindent\textbf{Superpoint refinement.}
We also utilize superpoint~\cite{superpoint} to complement $\hat{M}$. Based on the assumption all the points in a superpoint should belong to the same instance, we determine instance IDs of superpoints as the mode IDs within the points of their own. Then the instance IDs of superpoints are broadcast to all the belonging points, which enhances the completeness of instance masks by leveraging spatial priors deriving from superpoints.

\subsubsection{Self-training pipeline} 
At the start-up stage of the model training, especially for data-efficient learning, the perception models tend to predict wrong pseudo labels for a large amount of unlabeled data, which can encumber the model training.
To lay prior knowledge for $\Phi^t$, we train the model only with the labeled data for a few epochs as a warm-up.
In the warm-up, to leverage the available semantic and instance annotations, we employ the mentioned several losses for labeled data, which can be formulated as $\mathcal{L}^l = \mathcal{L}^l_{sem}+\mathcal{L}^l_{ins}$,
where $\mathcal{L}_{sem}$ and $\mathcal{L}_{ins}$ are defined in Eqn~.\ref{eq:loss_sem} and Eqn~.\ref{eq:loss_ins}, respectively.
Through learning from labeled data, InsTeacher3D builds basic instance discrimination ability for pseudo labeling. Also, $\mathcal{L}^l_{sem}$ can impart the model with the knowledge of category names.

Obtained the initialized $\Phi^t$ after the warm-up, we set up the $\Phi^s$ with a replica of $\Phi^t$, and the teacher is ready for the generation of instance pseudo labels cooperating with the dynamic mask generation module in Sec.~\ref{sec:dmg}. 
Then, having obtained high-quality instance pseudo labels $\hat{M}^u$, we can leverage instance consistency regularization upon the unlabeled data $P^u$ to boost the performance of $\Phi^t$ and $\Phi^s$. We introduce instance consistency regularization through localization, representation, and reconstruction terms, which can be formulated as:
\begin{equation}\label{eq:consistency-regularization}
    \mathcal{L}^u = \mathcal{L}^u_{ins} = \mathcal{L}^u_{loc}+\mathcal{L}^u_{rep}+\mathcal{L}^u_{rec}.
\end{equation}
All the above losses will be defined in Sec.~\ref{sec:instance-segmentation}.

Therefore, after the warm-up stage, the final training loss for $\Phi^s$ consists of supervised loss and consistency loss, which can be expressed as:
\begin{equation}
    \mathcal{L} = \mathcal{L}^l + \mathcal{L}^u = \mathcal{L}^l_{sem} + \mathcal{L}^l_{ins} + \mathcal{L}^u_{ins}.
\end{equation}
Meanwhile, as stated in Sec.~\ref{sec:mt}, the teacher model $\Phi^t$ is updated via EMA according to Eqn~.\ref{eq:ema}.

\section{Revisiting DKNet in Instance Consistency Regularization} \label{sec:dknet}
We argue that noisy semantic pseudo labels can cause ambiguity in model learning and the instance pseudo labels are more reliable, which prompts us to propose instance consistency regularization to fully exploit instance information and get rid of the reliance on noisy semantic prediction.
Through a series of experiments in Sec.~\ref{serial-parallel}, we prove that DKNet plays a crucial role in instance consistency regularization framework. In this section, we briefly review the architecture of DKNet and revisit the priorities of DKNet in instance consistency regularization. 

As depicted in Fig.~\ref{fig:pipeline-dknet}, the pivot of DKNet is to encode every instance into a discriminative representation, \textit{instance kernel}, which can recover the instance mask via scanning the whole scene.
Specifically, DKNet encodes instances into instance kernels via centroid localization and representation rather than relying on semantic prediction; and reconstructs the instance masks by dynamic convolution. \textit{This paradigm performs instance segmentation parallel with the semantic segmentation branch and encourages the model to explore informative instance discrimination instead of relying on noisy semantic segmentation.}
In addition, as previous works have shown~\cite{votenet,gicn,rpgn,liu2022weakly}, the areas of object centroids provide rich instance context and can yield purer segmentation results. Most of the over-segmentation and under-segmentation appear in the border of instances, while the center parts tend to be united. \textit{DKNet focuses on the localization and representation of instance centroids effectively excluding numerous noises for instance discrimination.} Subsequently, we discuss how these priorities are manifested in the detailed design of DKNet. 

\subsection{Feature extraction.}
Given $N$ points with colors (r, g, b) and coordinates (x, y, z) as the raw point cloud $P \in \mathbb{R}^{N\times6}$, the base instance segmentation model should produce point-wise semantic categories $S = \{S_n\in\{1,2,\cdots, C\}\}_{n=1}^N$ and soft instance masks $R \in \mathbb{R}^{N\times I}$.
$C$ is a fixed value for every dataset denoting the object category number, while the instance number $I$ varies in different scenes. 
We can determine the class of every instance via category voting among belonging points.
Beginning with the raw point cloud $P$, the feature extractor adopts a sparse U-Net~\cite{unet,ssc} following the previous works~\cite{pointgroup,dknet}, which produces the point cloud features. Obtained point cloud features, several MLPs further processed the features to predict point-wise semantic scores $B\in \mathbb{R}^{N\times C}$, centroid offsets $O\in \mathbb{R}^{N\times 3}$, and centroid heatmap $H=\{H_n \in [0,1]\}_{n=1}^N$.

\subsection{Semantic segmentation.}
 $S$ can be easily obtained by performing a channel-wise maximum upon $B$ to realize semantic segmentation. To learn knowledge of the name of semantic categories from labeled data, we employ cross entropy loss and multi-class dice loss for supervising semantic scores $B$. The semantic loss $\mathcal{L}_{sem}$ can be formulated as:
\begin{equation}\label{eq:loss_sem}
\begin{aligned}
\mathcal{L}_{sem} = &-\frac{1}{N} \sum_{n=1}^{N} {CE(B_n, \hat{B}_n)}\\
&+\frac{1}{C}\sum_{c=1}^C(1 - 2\frac{\sum_{n=1}^N B_{n,c}\hat{B}_{n,c}}{\sum_{n=1}^N B_{n,c}^2+\sum_{n=1}^N \hat{B}_{n,c}^2})\\
\end{aligned},\\
\end{equation}
where the one-hot vector $\hat{B}_n$ is the ground truth semantic score, $CE$ denotes the cross-entropy loss, and the second term is a multi-class dice loss.

\subsection{Instance segmentation.} \label{sec:instance-segmentation}
 To exhaustively explore instance discrimination, we introduce a localize-then-aggregate paradigm to encode instance kernels and a dynamic convolution~\cite{dfn} module to decode instance masks. Thus, the segmentation procedure can be divided into three stages: instance localization, representation, and mask reconstruction. The instance labels are given in the form of $\hat{M} = \{\hat{M}_n\in\{-1,1,2,\cdots, I\}\}_{n=1}^N$, each element $\hat{M}_n$ of which is the instance ID of point $P_n$. $-1$ means background or unlabeled point.
 
 \noindent \textbf{Instance localization.} To avoid the impact of the potential noise, we first localize instance centroids in a coarse-to-fine manner, where we predict coarse $O$ and further build accurate heatmaps $H$.
 The instance encoder localizes objects based on $H$, each element of which is a score in $[0,1]$ reflecting how close each point is to its respective instance centroid. As centroids are expected to have the highest value among their neighbors, an iterative instance searching algorithm~\cite{dknet} is applied to localize local maxima in $H$ as instance candidates. 
 We supervise the procedure of instance localization focusing on instance centroids. For $i^{th}$ instance, its instance centroid $\hat{C}_i$ is the mean coordinates of points whose instance IDs are $i$ in $\hat{M}$. If $n^{th}$ points belonging $i^{th}$ instance, its ground truth $\hat{O}_n$ and $\hat{H}_n$ can be formulated as:
\begin{equation}
    \hat{O}_n = \hat{C}_i - X_n ,\ \hat{H}_n = exp(-\Vert\hat{O}_n\Vert^2/ L_i^2),
\end{equation}
where $X_n$ denotes the coordinate of $n^{th}$ point and $L_i$ is a coefficient depends on the size of $i^{th}$ instance. The localization loss $\mathcal{L}_{loc}$ supervises the length and direction of offsets and the centroid heatmap, which can be formulated as:
\begin{equation}
\begin{aligned}
    \mathcal{L}_{loc}=&\frac{1}{N'} \sum_{n=1}^{N} (\Vert O_n -\hat{O}_n\Vert + \frac{O_n \cdot \hat{O}_n}{\Vert O_n \Vert \cdot \Vert \hat{O}_n \Vert}+\\
    &\vert H_n - \hat{H}_{n}\vert) \cdot \mathbbm{1}(P_n), 
\end{aligned}
\end{equation}
where $\mathbbm{1}(P_n)$ is an indicator function that outputs 1 when $n^{th}$ point is foreground, otherwise outputs $0$. $N'$ is the number of foreground points in every scene, which can be described as $N'=\sum^N_{n=1}\mathbbm{1}(P_n)$. Note that $H$ remains unchanged in the following processing.

 \noindent \textbf{Instance representation.} Obtained $Q$ candidates via centroid localization, we collect the spherical neighbors of the centroid candidates to represent these candidates. Before building instance kernels to recover the masks, it is expected that the instances and kernels are in one-to-one correspondence. However, it is inevitable to generate redundant candidates for some large instances. Hence, a learning-based candidate affinity measure module produces instance affinity matrix $A\in \mathbb{R}^{Q\times Q}$, based on which a greedy aggregation algorithm~\cite{dknet} further merges candidate features into $I$ instance kernels. We expect that each element $A_{j,k}$ should reflect, in the embedding domain, how similar candidates $j$ and $k$ are. 
Therefore, we set the ground truth $\hat{A}_{j,k}$ as $1$ if $\hat{M}_j$ equal to $\hat{M}_k$, and vice versa. 
To supervise the learning of $A$, we apply a binary cross-entropy (BCE) loss as:
\begin{align}
    \mathcal{L}_{rep} = BCE(A, \hat{A}).
\end{align}
This loss $\mathcal{L}_{rep}$ can facilitate the discriminative representation of instances in a contrastive manner~\cite{contrast,pointcontrast,pan2023find} by pushing candidates of different objects away and pulling the candidates of the same objects close. In addition, for each merged instance, among the belonging candidates, we select the candidate with the highest score in $H$ as its centroid.

\noindent \textbf{Instance reconstruction.} Once the kernels are built, we can simply scan the whole scene with dynamic kernels~\cite{dfn,dyco3d,dknet,isbnet} to reconstruct the soft instance masks $R$. The key to precise mask reconstruction is to build accurate matches between mask prediction and ground truth.
We use the distances between centroids of merged instances and instance ground truths as matching costs, which are then processed by the Hungarian algorithm~\cite{hungarian} to build one-to-one training pairs. Compared with the commonly used matching cost, Intersection-over-Union~(IoU) between predictions and ground truth masks, the centroid distances help establish the correct matches even with errors in the border of instance masks in the pseudo labels.  
Afterward, the mask reconstruction is supervised by a BCE loss and a dice loss~\cite{dice}, which can further overcome the class imbalance. 
The instance mask targets $\hat{R} \in \mathbb{R}^{N\times I}$ can be easily obtained by converting $\hat{M}$ to one-hot masks.
Thus, the reconstruction loss $\mathcal{L}_{rec}$ can be formulated as:
\begin{equation}
\begin{aligned}
    \label{eq: mask-generation-loss}
    \mathcal{L}_{rec} = &\frac{1}{I'}\sum_{i=1}^{I} (BCE({R^T}_i, {\hat{R^T}}_i)+ 
    \\&1- 2\frac{{R^T}_i\cdot{\hat{R^T}}_i}{\vert {R^T}_i\vert+\vert {\hat{R^T}}_i\vert})\cdot \mathbbm{1}(iou_i>0.5)\,,
\end{aligned}
\end{equation}
where $iou_i$ denotes the IoU between ${R^T}_i$ and $\hat{R^T}_i$, and $\mathbbm{1}$ is an indicator function.

In conclusion, the learning procedure of instance segmentation can be formulated as:
\begin{equation}\label{eq:loss_ins}
    \begin{aligned}
    \mathcal{L}_{ins} = &\mathcal{L}_{ins}(O,H,A, R, \hat{M}) \\
    = &\mathcal{L}_{loc}(O, H, \hat{O}, \hat{H})+\mathcal{L}_{rep}(A, \hat{A})\\
     &+\mathcal{L}_{rec}(R, \hat{R}).
    \end{aligned} 
\end{equation}

We strongly recommend readers refer to DKNet~\cite{dknet} for more details on the candidate localization, aggregation, and instance reconstruction procedures.

\begin{table*}[!t]
\begin{center}
\caption{\textbf{Quantitative results on the ScanNetV2 validation set and hidden test set on the online benchmark.} We report the results under four annotation rates. The performance improvements labeled as ``$\Delta$ $\uparrow$'' are obtained by comparing semi-supervised models with the corresponding ``Sup.-only'' models.}
\vspace{-0.4cm}
\label{table:performance_scan}
\resizebox{\linewidth}{!}{
\begin{tabular}{@{}l|l|lll|lll|lll|lll@{}}
\toprule
\multirow{2}*{Set}&\multirow{2}*{Approach}&\multicolumn{3}{c|}{$1\%$}&\multicolumn{3}{c|}{$5\%$}&\multicolumn{3}{c|}{$10\%$}&\multicolumn{3}{c}{$20\%$}\\
~&~&mAP&$AP_{50}$&$AP_{25}$&mAP&$AP_{50}$&$AP_{25}$&mAP&$AP_{50}$&$AP_{25}$&mAP&$AP_{50}$&$AP_{25}$ \\
\midrule
\multirow{8}*{Val}&\cellcolor{yellow!12.5}\textit{Sup.-only (Ours)}&\cellcolor{yellow!12.5}2.1&\cellcolor{yellow!12.5}4.9 &\cellcolor{yellow!12.5}10.5 &\cellcolor{yellow!12.5}18.8 &\cellcolor{yellow!12.5}31.6 &\cellcolor{yellow!12.5}42.3 &\cellcolor{yellow!12.5}30.1 &\cellcolor{yellow!12.5}46.1 &\cellcolor{yellow!12.5}58.5 &\cellcolor{yellow!12.5}35.3 &\cellcolor{yellow!12.5}51.0&\cellcolor{yellow!12.5}63.4 \\
~&\cellcolor{yellow!12.5}\textit{Sup.-only (TWIST)}&\cellcolor{yellow!12.5}5.1&\cellcolor{yellow!12.5}9.8 &\cellcolor{yellow!12.5}17.6 &\cellcolor{yellow!12.5}18.2 &\cellcolor{yellow!12.5}32.0 &\cellcolor{yellow!12.5}47.0 &\cellcolor{yellow!12.5}26.7 &\cellcolor{yellow!12.5}42.8 &\cellcolor{yellow!12.5}58.9 &\cellcolor{yellow!12.5}29.3 &\cellcolor{yellow!12.5}47.9&\cellcolor{yellow!12.5}63.0 \\
~&PointContrast~\cite{pointcontrast}&7.2&12.5 &20.3 &19.4 &35.4 &48.5 &27.0 &43.9 &59.5 &30.2 &49.5&63.6 \\
~&CSC~\cite{csc}&7.1&13.0 &21.2 &20.9 &36.7 &50.6 &27.3 &45.0 &60.2 &30.6 &50.3&64.1 \\
~&WS3D~\cite{liu2022weakly}&-&\textbf{32.5}&-&-&\textbf{45.6}&-&-&49.2&-&-&51.3&-\\
~&TWIST~\cite{twist}&9.6 &17.1 &\textbf{26.2} &27.0 &44.1&\textbf{56.2} &30.6 &49.7 &63.0 &32.8 &52.9&66.8 \\
~&TWIST\ $\Delta$ $\uparrow$ &\textcolor{brown}{\scriptsize +4.5  } &\textcolor{brown}{\scriptsize +7.3  } &\textcolor{brown}{\scriptsize +8.6  } &\textcolor{brown}{\scriptsize +8.8  } &\textcolor{brown}{\scriptsize +12.1  } & \textcolor{brown}{\scriptsize +9.2  } &\textcolor{brown}{\scriptsize +3.9  } & \textcolor{brown}{\scriptsize +6.9  } & \textcolor{brown}{\scriptsize +4.1  } &\textcolor{brown}{\scriptsize +3.5  } &\textcolor{brown}{\scriptsize +5.0  }&\textcolor{brown}{\scriptsize +3.8  } \\
~&\textbf{Ours}&\textbf{10.8} &16.6 &21.2 & \textbf{30.0} &44.7 &55.6 &\textbf{39.6} &\textbf{56.9} &\textbf{67.0} &\textbf{42.7} &\textbf{58.3} &\textbf{68.2} \\
~&\textbf{Ours}\quad\ $\Delta$ $\uparrow$ &\textcolor{brown}{\textbf{\scriptsize +8.7}}  &\textcolor{brown}{\textbf{\scriptsize +11.7}} &\textcolor{brown}{\textbf{\scriptsize +10.7}} & \textcolor{brown}{\textbf{\scriptsize +11.2}} &\textcolor{brown}{\textbf{\scriptsize +13.1}}&\textcolor{brown}{\textbf{\scriptsize +13.3}} &\textcolor{brown}{\textbf{\scriptsize +9.8} }&\textcolor{brown}{\textbf{\scriptsize +10.8}}&\textcolor{brown}{\textbf{\scriptsize +8.5}} & \textcolor{brown}{\textbf{\scriptsize +7.4}} &\textcolor{brown}{\textbf{\scriptsize +7.2}} &\textcolor{brown}{\textbf{\scriptsize +4.8}} \\
~&\cellcolor{gray!12.5}\textit{TWIST~\cite{twist}+CSC~\cite{csc}}&\cellcolor{gray!12.5}11.5&\cellcolor{gray!12.5}20.0 &\cellcolor{gray!12.5}31.1 &\cellcolor{gray!12.5}28.6 &\cellcolor{gray!12.5}45.9 &\cellcolor{gray!12.5}58.2 &\cellcolor{gray!12.5}32.8 &\cellcolor{gray!12.5}51.5 &\cellcolor{gray!12.5}65.1 &\cellcolor{gray!12.5}34.1 &\cellcolor{gray!12.5}53.7&\cellcolor{gray!12.5}67.8 \\
~&\cellcolor{gray!12.5}\textit{Ours+CSC~\cite{csc}}&\cellcolor{gray!12.5}13.7&\cellcolor{gray!12.5}21.5 &\cellcolor{gray!12.5}29.3 &\cellcolor{gray!12.5}28.3 &\cellcolor{gray!12.5}42.9 &\cellcolor{gray!12.5}55.3 &\cellcolor{gray!12.5}38.3 &\cellcolor{gray!12.5}54.7 &\cellcolor{gray!12.5}66.5 &\cellcolor{gray!12.5}41.8 &\cellcolor{gray!12.5}56.9&\cellcolor{gray!12.5}67.5 \\
\midrule
\multirow{5}*{Test}&PointContrast~\cite{pointcontrast}&5.7 &11.9 &20.6 &17.6 &29.8 &43.8 &25.9 &43.2 &57.3 &30.4 &48.8 &67.6\\
~&CSC~\cite{csc}&5.6 &11.7 &21.7 &19.6 &32.5 &49.2 &26.4 &44.0 &61.5 &32.2 &52.9 &68.3\\
~&TWIST~\cite{twist}+CSC~\cite{csc}&10.8 &\textbf{18.6} &\textbf{28.8} &25.7 &{42.1} &\textbf{59.2} &{29.5} &{48.1} &{66.9} &34.2 &{55.0}&{69.3} \\
~&\textbf{Ours}&\textbf{11.7}&{16.2} &20.3 &\textbf{31.0} &\textbf{44.3} &56.3 &\textbf{41.6} & \textbf{59.8}&\textbf{73.8}&\textbf{47.3} &\textbf{64.6} &\textbf{73.5} \\
\bottomrule
\end{tabular}}
\end{center}
\vspace{-0.6cm}
\end{table*}

\section{Experiments}
In this section, we first evaluate InsTeacher3D and compare it with previous state-of-the-art unsupervised and semi-supervised 3D instance segmentation approaches. Then, a series of ablation experiments are designed to verify the components of InsTeacher3D.
We build experiments on two popular indoor 3D point cloud datasets: ScanNetV2~\cite{scannet} and S3DIS~\cite{s3dis}, and one large-scale outdoor dataset, STPLS3D~\cite{chen2022stpls3d}.
\subsection{Implementation Details}
\subsubsection{Datasets}
All three datasets, ScanNetV2, S3DIS, and STPLS3D provide dense instance and semantic labels.
ScanNetV2 consists of $1613$ scenes in total, and the scenes are split into subsets with $1201$, $312$, and $100$ scans for training, validation, and testing, respectively. This dataset includes $18$ object classes and $2$ invalid classes. S3DIS has $6$ areas containing $272$ rooms with $13$ object categories. Following the common split, we use the Area-$5$ as the validation set and the others as the training set. STPLS3D is a synthetic large-scale outdoor point cloud dataset, where $25$ urban scenes totaling $6 km^2$ are densely labeled with $14$ classes. We follow the common split~\cite{chen2022stpls3d,softgroup} to obtain the train/validation set. 
Across three datasets, we employ mean average precision (mAP) under different IoU thresholds as evaluation metrics. 

\subsubsection{Data-Efficient Settings}
We follow the limited reconstruction setting in previous works~\cite{twist} to construct training sets, where the annotation rates are in $\{1\%, 5\%, 10\%, 20\%\}$. On ScanNetV2, we directly adopt the training scan splits from the Data-Efficient ScanNetV2 benchmark~\cite{csc}. Meanwhile, we also split S3DIS and STPLS3D in the same four ratios by randomly sampling from the complete training set. For robust self-training, we utilize random flipping and rotation as weak augmentation, while the strong augmentation further implements point-wise jitters, elastic, and color distortion. 

\subsubsection{Semi-Supervised Learning Details}
We train the whole semi-supervised network on a single RTX $3090$ GPU and adopt an Adam~\cite{adam} optimizer with an initial learning rate of $0.001$. The learning rate is subsequently adjusted by a cosine scheduler~\cite{cosine}. We voxelize the point clouds with the size of $2cm$ for both ScanNetV2 and S3DIS. Besides, for each scene in S3DIS, the points are randomly sub-sampled to ${\sim}1/4$ to reduce point density. 
We train the semi-supervised network for $400$ and $600$ epochs in ScanNetV2 and S3DIS, where the first $96$ and $160$ epochs are reserved for a warm-up, respectively. In the remaining epochs, we train $\Phi^s$ with resampled batches of batch size $4$, which consist of $2$ labeled scenes and $2$ unlabeled scenes. Compared to $\Phi^t$ updated after every step, the pseudo labels are generated after every epoch. To obtain high-precision instance pseudo labels, we remove the instances with confidence scores lower than $0.5$ and instances occupying less than $100$ points. The scores are evaluated as the average scores of foreground points in instance and semantic masks. As S3DIS and STPLS3D do not provide superpoints, we deactivate the superpoint refinement module in the implementation of these datasets.

\subsection{Main Results}

\subsubsection{Performance on indoor datasets}
We evaluate the semi-supervised 3D instance segmentation performance of our model on both ScanNetV2 validation and hidden test sets in Table~\ref{table:performance_scan}.
The performances on the S3DIS validation set are also reported in Table~\ref{table:performance_scan_s3dis}.

\begin{figure*}[!t]
\centering
\includegraphics[width=1\linewidth]{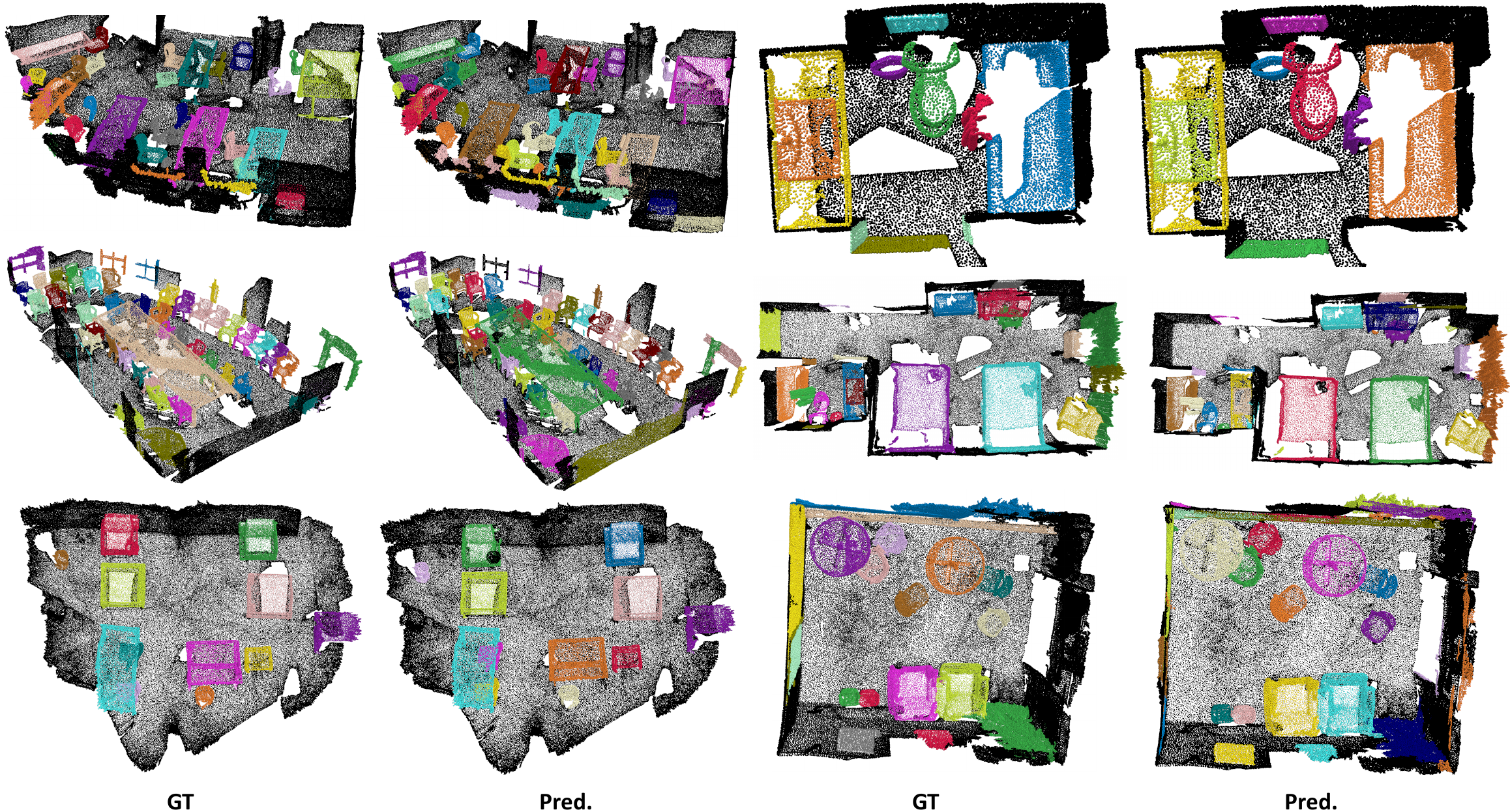}
\caption{\textbf{More visualization results of instance segmentation on ScanNetV2 validation set.} 'Pred.' means the prediction of InsTeacher3D with only $20\%$ data labeled. Best viewed in color.}
\label{fig:visualization}
\vspace{-0.4cm}
\end{figure*}

\noindent \textbf{Superiority in instance discrimination.} On the one hand, InsTeacher3D significantly boosts the baseline model (indicated as ``Sup.-only'' in the above tables) via instance consistency regularization for unlabeled data.
On the other hand, compared with the previous state of the arts, our approach outperforms other unsupervised and semi-supervised approaches in most metrics under most settings.  
To be specific, given $20\%$ labeled scenes, our approach achieves $42.7\%$ mAP on the ScanNetV2 validation set, which surpasses TWIST~\cite{twist} by $9.9\%$ and relatively boosts ``Sup.-only'' model by $21.0\%$. 
We also compare our approach trained with only $20\%$ labels with fully supervised approaches in Table~\ref{table:comparison}, where InsTeacher3D shows competitive performance.
In addition, we present visual results of instance segmentation on ScanNetV2 validation set in Fig.~\ref{fig:visualization} as a qualitative evaluation. Our approach produces accurate instance segmentation results. 

The high-quality instance pseudo labels in consistency regularization become of utmost importance for building strong instance discrimination, which is in turn the sine qua non for successful instance segmentation. Since higher IoU thresholds in mAP metrics provide a more accurate measure of the ability of approaches to capture instance shapes, the extraordinary performance of InsTeacher3D under higher thresholds (mAP) proves the effectiveness of our approach in improving instance discrimination. This superiority further enables it to demonstrate the best performance in most settings, especially with high annotation rates.

Besides, though DKNet~\cite{dknet} only obtains poor performance under the extremely limited setting, InsTeacher3D effectively boosts it to a comparable level with previous approaches and consistently obtains higher improvements compared with TWIST.  

\noindent \textbf{Inferiority under low label rates.} One possible reason for the inferior performance of DKNet under the $1\%$ setting is the learning of approaches in kernel-based paradigm relies on numerous instance-level matches while the few instances in limited scenes, $12$ in ScanNetV2 and only $2$ in S3DIS, cannot effectively support the establishment of instance discrimination. 
Due to the same reason, InsTeacher3D only shows comparable performance in S3DIS with TWIST. 

The previous work~\cite{twist} inspires us to incorporate unsupervised techniques to alleviate the noisy learning procedure, especially for low label rates. Thus, we combine InsTeacher3D and CSC~\cite{csc} for further performance improvement. As shown in Table~\ref{table:performance_scan} and Table~\ref{table:performance_scan_s3dis}, the results highlighted with gray color demonstrate that InsTeacher3D can be further enhanced when integrated with CSC under low label rates. Notably, CSC significantly improves the performances of InsTeacher3D under the $1\%$ setting in ScanNetV2 and all settings in S3DIS. This also partly substantiates that the inferior performance of InTeacher3D under low annotation rates is attributed to its lack of robust instance discrimination. However, under high annotation rates, we notice slight performance degradation in Table~\ref{table:performance_scan} ($5\%$, $10\%$, $20\%$ settings). We analyze that the degradation derives from the confusing knowledge from the loose constraints in unsupervised learning, which may hinder the instance consistency regularization under high label rates.

\subsubsection{Performance on outdoor datasets}
We evaluate DKNet and InsTeacher3D on large-scale outdoor dataset, STPLS3D~\cite{chen2022stpls3d} validation set in Table~\ref{table:performance_scan_stpls3d}. DKNet shows a convincing performance ($45.5$ mAP) on the eval set, demonstrating its scalability and robustness on large-scale outdoor datasets. Besides, as this work mainly focuses on semi-supervised 3D instance segmentation, we report the performance of InsTeacher3D on STPLS3D eval set, where InsTeacher3D shows surprising performance under four semi-supervised settings. Specifically, trained with only $20\%$ labeled data, InsTeacher3D shows $93.4\%$ performance ($42.5$ mAP) of the fully-supervised DKNet ($45.5$ mAP). Besides, trained with only $1\%$ data, InsTeacher3D achieves $26.5$ mAP and surpasses the fully-supervised approach, PointGroup~\cite{pointgroup}. The above results prove that InsTeacher successfully exploits the potential instance knowledge from unlabeled data.
In conclusion, InsTeacher3D and DKNet are robust to diverse environments, and InsTeacher3D provides a strong baseline for semi-supervised 3D instance segmentation on outdoor scenes.

\begin{table}[!t]
\small
\begin{center}
\caption{\textbf{Quantitative results on S3DIS validation set Area-$\bf{5}$.} mAP is reported.}
\label{table:performance_scan_s3dis}
\vspace{-0.4cm}
\begin{tabular}{@{}l|cccc@{}}
\toprule
Approach&$1\%$&$5\%$&$10\%$&$20\%$\\
\midrule
Sup-only&$9.5$&$17.4$ &$26.4$ &$33.0$\\
PointContrast~\cite{pointcontrast}&$13.4$&$22.9$ &$27.1$ &$31.2$\\
CSC~\cite{csc}&$14.6$&$24.9$ &$29.7$ &$33.5$\\
TWIST~\cite{twist}&$\textbf{17.9}$&$27.1$ &$\textbf{33.6}$ &$36.7$\\
\textbf{Ours}& $16.9$&$\textbf{27.9}$ &${31.8}$ &$\textbf{39.1}$ \\
\cellcolor{gray!12.5}\textit{TWIST~\cite{twist}+CSC~\cite{csc}}&\cellcolor{gray!12.5}18.9&\cellcolor{gray!12.5}29.3 &\cellcolor{gray!12.5}35.0&\cellcolor{gray!12.5}37.9\\
\cellcolor{gray!12.5}\textit{Ours+CSC~\cite{csc}}&\cellcolor{gray!12.5}17.4&\cellcolor{gray!12.5}29.8&\cellcolor{gray!12.5}35.2&\cellcolor{gray!12.5}43.9\\
\bottomrule
\end{tabular}
\end{center}
\vspace{-0.8cm}
\end{table}
\begin{table}[!t]
\small
\begin{center}
\caption{\textbf{Comparison with different semi-supervised and fully supervised instance segmentation approaches.} Trained with $20\%$ annotation, InsTeacher3D demonstrates comparable performance with fully supervised approaches.}
\label{table:comparison}
\vspace{-0.4cm}
\begin{tabular*}{0.9\linewidth}{@{}@{\extracolsep{\fill}}l|l|ccc@{}}
\toprule
Rate& Approach &mAP&$AP_{50}$&$AP_{25}$\\
\midrule
\multirow{2}*{$20\%$}&TWIST~\cite{twist}&32.8&52.9 &66.8  \\
~&\textbf{InsTeacher3D}&42.7&58.3 &68.2 \\
\midrule
\multirow{6}*{$100\%$}&GSPN~\cite{gspn}&19.3&37.8&53.4\\
~&PointGroup~\cite{pointgroup}&34.8&56.9&71.3\\
~&3D-MPA~\cite{3dmpa}&35.3&59.1&72.4\\
~&DyCo3D~\cite{dyco3d}&35.4&57.6&-\\
~&HAIS~\cite{hais}&43.5&64.1&75.6\\
~&\textbf{DKNet}&50.8&66.7&76.9 \\
\bottomrule
\end{tabular*}
\end{center}
\vspace{-0.6cm}
\end{table}
\begin{table}[!t]
\small
\begin{center}
\caption{\textbf{Quantitative results on STPLS3D validation set under fully- and semi-supervised settings.} }
\label{table:performance_scan_stpls3d}
\vspace{-0.4cm}
\begin{tabular*}{\linewidth}{@{}@{\extracolsep{\fill}}l|c|l|ccc@{}}
\toprule
Setting&Rate&Approach&mAP&$AP_{50}$&$AP_{25}$\\
\midrule
\multirow{4}*{Fully-}&\multirow{4}*{$100\%$}&PointGroup~\cite{pointgroup}&$23.3$&$38.5$ &$48.6$ \\
~&~&HAIS~\cite{hais}&$35.1$&$46.7$ &$52.8$ \\
~&~&SoftGroup~\cite{softgroup}&$46.2$&$61.8$ &$69.4$\\
~&~&\textbf{DKNet}& $45.5$&$58.9$ &${64.0}$  \\
\noalign{\smallskip}
\hline \hline
\noalign{\smallskip}
\multirow{4}*{Semi-}&$20\%$&\multirow{4}*{\textbf{InsTeacher3D}}&$42.5$&$54.8$&$60.3$\\
~&$10\%$&~&$34.5$&$44.0$&$48.7$\\
~&$5\%$&~&$30.1$&$38.0$&$41.8$\\
~&$1\%$&~&$26.5$&$33.8$&$37.4$\\
\bottomrule
\end{tabular*}
\end{center}
\vspace{-0.6cm}
\end{table}

\begin{figure*}[!t]
\centering
\subfigure[Evaluation of instance pseudo labels.]{\includegraphics[width=0.48\linewidth]{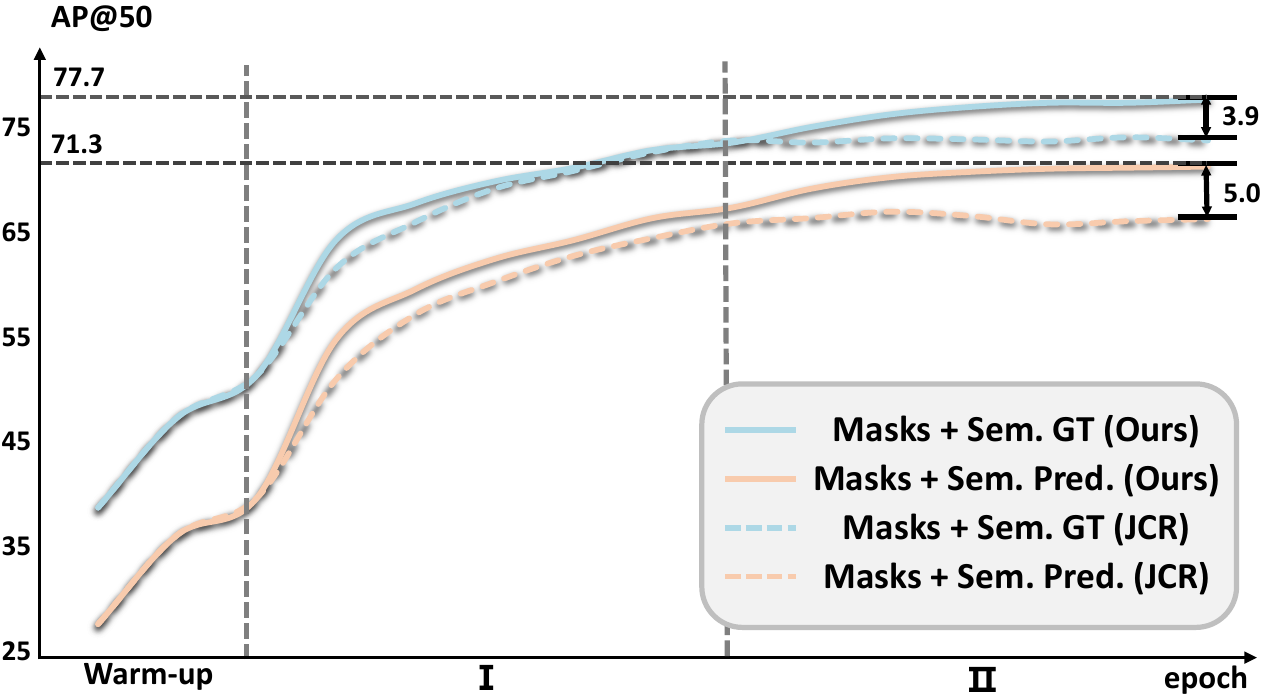}}\hfill
\subfigure[Evaluation of semantic pseudo labels.]{\includegraphics[width=0.48\linewidth]{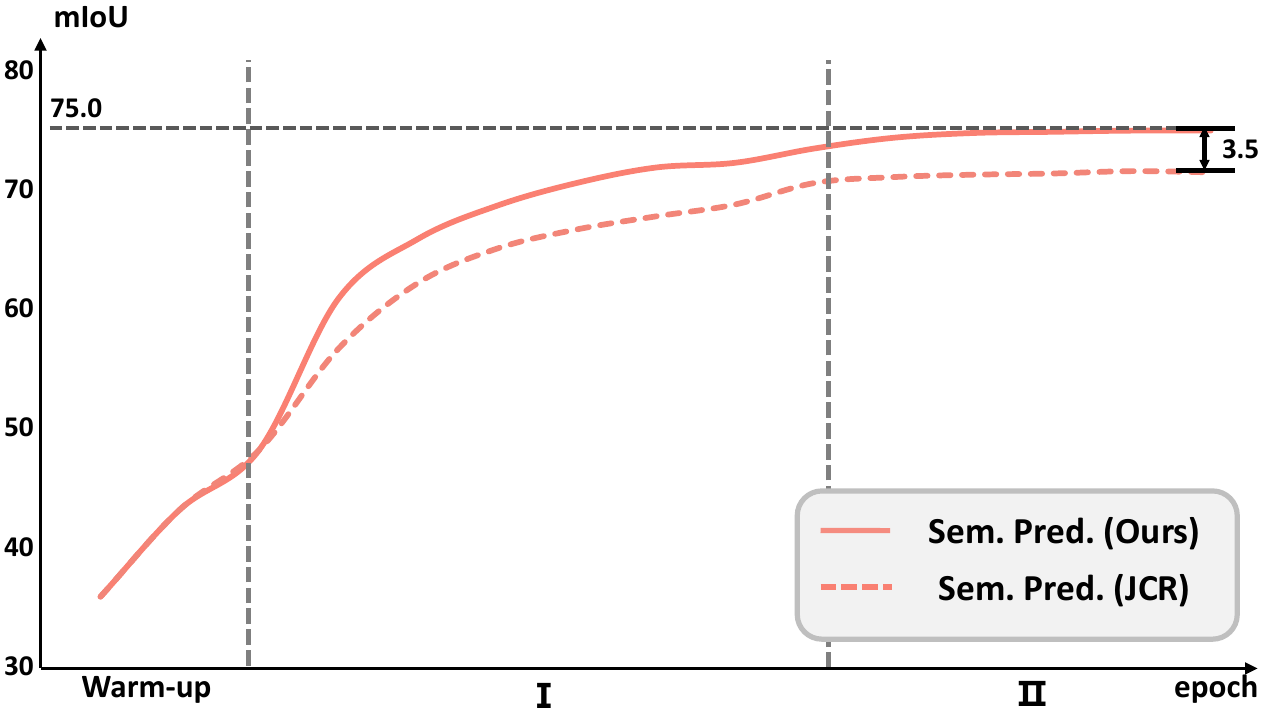}}
\vspace{-0.3cm}
\caption{\textbf{Evaluations of the quality of instance and semantic pseudo labels.} We report AP@50 and mIoU for instance and semantic segmentation on the ScanNetV2 training set, respectively. ``Sem. GT'' means we combine the semantic ground truth with instance mask predictions for evaluation, while we use category predictions in ``Sem. Pred.''. ``JCR'' is joint consistency regularization. The qualities of pseudo labels from 64 to 400 epochs are recorded in the table, and, for further analysis, we divide the entire training procedure into three stages: warm-up, stage \uppercase\expandafter{\romannumeral1}, and stage \uppercase\expandafter{\romannumeral2}.}
\label{fig:pseudo-label}
\vspace{-0.4cm}
\end{figure*}

\subsubsection{Evaluation of pseudo labels.}
As illustrated in Fig.~\ref{fig:pseudo-label}, we report the qualities of instance and semantic pseudo labels on ScanNetV2 training set over epochs. For evaluation for instance pseudo labels in Fig.~\ref{fig:pseudo-label}(a), AP@50 are calculated under two settings. We first combine instance masks with semantic segmentation prediction as ``Masks +Sem. Pred.'', which is influenced by both instance and semantic factors. In contrast, assigning true categories for all masks, ``Masks +Sem. GT'' accurately reflects the quality of instance pseudo labels and obtains a $6.4\%$ performance improvement compared with ``Masks +Sem. Pred.''. Besides, if joint instance and semantic consistency are employed in consistency regularization, all the metrics exhibit a noticeable decline.

In addition, for further analysis, we divide the entire training procedure into three stages: warm-up, stage \uppercase\expandafter{\romannumeral1}, and stage \uppercase\expandafter{\romannumeral2}. In the warm-up stage, the segmentation model learns knowledge for instance segmentation from all available labeled data, which ensures the basic quality of initial pseudo labels. Then, in the following stages, consistency regularization is introduced to leverage unlabeled data. As shown in Fig.~\ref{fig:pseudo-label}, the quality of pseudo labels is effectively improved in the stage \uppercase\expandafter{\romannumeral1}. The performance of the self-training semi-supervised segmentation model heavily depends on the quality of pseudo labels. We find that learning less but accurate knowledge is more helpful than learning more but noisy knowledge. At the end of the stage \uppercase\expandafter{\romannumeral1}, the models trained with joint consistency tend to converge and only generate inferior pseudo labels. In contrast, models with our self-training strategy focusing on instance consistency produce pseudo labels with increasing qualities in stage \uppercase\expandafter{\romannumeral2}.

\subsection{Ablation Study}
To evaluate the effectiveness of components of InsTeacher3D, we conduct a series of ablation experiments on the ScanNetV2 validation set under $20\%$ annotation rate. 

\subsubsection{Ablation on different components}
In this work, we aim to use the power of instance consistency and mitigate the impact of noisy semantic prediction in semi-supervised 3D instance segmentation. To this end, we integrate parallel segmentation model and instance consistency regularization framework into our InsTeacher3D. As shown in Table~\ref{table:effectiveness}, we conduct ablation to evaluate the effectiveness of each component. 

\noindent \textbf{Effect of mean teacher framework.}
TWIST~\cite{twist} only employs one model to generate pseudo labels for unlabeled data and subsequently learn from the noisy pseudo labels. In contrast, the mean teacher~\cite{mt} framework adopts a self-ensembling teacher model to generate high-quality pseudo labels and utilize different augmentations to leverage consistency regularization. Group \romannumeral2\  with mean teacher significantly boosts the self-training framework used in TWIST (group \romannumeral1) by $2.0\%$ in AP@50. 

\noindent \textbf{Effect of DKNet.} Compared with serial segmentation models, the parallel model (DKNet) separates the mask reconstruction with noisy semantic segmentation, which enables the generation of high-quality pseudo labels. Thus, DKNet in group \romannumeral4-\romannumeral6\ effectively improve \romannumeral1-\romannumeral3\  with serial models by $3.8\%$, $2.4\%$, and $2.6\%$ AP@50, respectively. Besides, thanks to accurate instance pseudo labels, group \romannumeral4\ with an inferior self-training framework achieves comparable performance with group \romannumeral5.

\noindent \textbf{Effect of instance consistency regularization.} In addition, we argue noisy semantic pseudo labels are still harmful to consistency regularization. Thus, we only employ instance consistency to learn from the numerous unlabeled data in group \romannumeral6, which outperforms the model with joint consistency regularization by $1.0\%$ AP@50 and overall boost the baseline (group \romannumeral1) by $9.9\%$ mAP.

\subsubsection{Evaluation of consistency regularization frameworks} \label{sec:framework}
We argue that the noisy semantic pseudo labels will lead to severe confirmation bias and design an ablation experiment to evaluate different consistency regularization frameworks shown in Table~\ref{table:self-training}. The sup-only model is trained with only supervised loss on labeled data. 
In semi-supervised segmentation tasks, joint consistency regularization is the most intuitive self-training strategy generating both instance and semantic pseudo labels to leverage all the available information, while semantic and instance consistency regularization frameworks produce only either of pseudo labels.
For a fair comparison, we also introduce semantic mask enhancement techniques, \emph{e.g}. thresholding and mutual enhancement like that in TWIST~\cite{twist}, to improve the quality of semantic pseudo labels. 
We analyze the experiment results in two aspects:
1) The model leveraging unlabeled data via only instance consistency shows the best performance while using semantic consistency is the worst. Further compared with the joint consistency regularization framework, instance consistency abandons semantic pseudo labels but improves the segmentation performance.
The above phenomenons demonstrate that engaging noisy semantic pseudo labels can lead to performance deterioration.
\begin{table}[!t]
\begin{center}
\caption{Ablation on different components of InsTeacher3D. ``ST'' denotes the self-training framework. ``MT'' is the mean teacher. ``Arch.'' means the architecture of the base segmentation model. ``CR'' is the consistency regularization strategy, where ``JCR'' and ``ICR'' respectively denote joint and instance consistency regularization.}
\label{table:effectiveness}
\begin{tabular*}{\linewidth}{@{}@{\extracolsep{\fill}}c|ccc|ccc@{}}
\toprule
Group&ST&Arch.&CL&mAP&$AP_{50}$&$AP_{25}$\\
\midrule
\romannumeral1 &w/o MT & Serial & JCR & $32.8$&$52.9$ &$66.8$  \\
\romannumeral2 &w/ MT  &Serial & JCR &$39.0$&$54.9$ &$66.5$ \\
\romannumeral3  &w MT  &Serial &ICR &$39.6$&$55.7$ &$67.3$ \\
\romannumeral4  &w/o MT  &Parallel &JCR &$41.1$&$56.7$ &$67.6$ \\
\romannumeral5  &w/ MT  &Parallel & JCR&$41.8$&$57.3$ &$67.5$ \\
\romannumeral6  &w/ MT  &Parallel & ICR &$42.7$&$58.3$ &$68.2$ \\
\bottomrule
\end{tabular*}
\end{center}
\vspace{-0.7cm}
\end{table}
\begin{table}[!t]
\small
\begin{center}
\caption{Comparison of different consistency regularization frameworks. ``Sem.'' and ``Ins.'' indicate employing semantic and instance pseudo labels, respectively. ``JCR'', ``SCR'', and ``ICR'' denote joint, semantic, and instance consistency regularization, respectively. For instance segmentation, $mAP$, $AP_{50}$, and $AP_{25}$ are presented. We also report mIoU for semantic segmentation.}
\label{table:self-training}
\begin{tabular*}{\linewidth}{@{}@{\extracolsep{\fill}}c|cc|cccc@{}}
\toprule
{Group}&Sem.&Ins.&mAP&$AP_{50}$&$AP_{25}$&mIoU\\
\midrule
Sup.-only & & &$35.3$&$51.0$ &$63.4$&$61.3$   \\
\midrule
JCR  &\checkmark &\checkmark &$41.8$&$57.3$ &$67.5$&$62.1$ \\
SCR  &\checkmark & &$41.1$&$55.0$ &$66.3$&$59.7$ \\
ICR  & &\checkmark &$42.7$&$58.3$ &$68.2$&$65.4$ \\
\bottomrule
\end{tabular*}
\end{center}
\vspace{-0.4cm}
\end{table}
\begin{figure}[!t]
\centering
\setlength{\abovecaptionskip}{-3pt}
\includegraphics[width=1\linewidth]{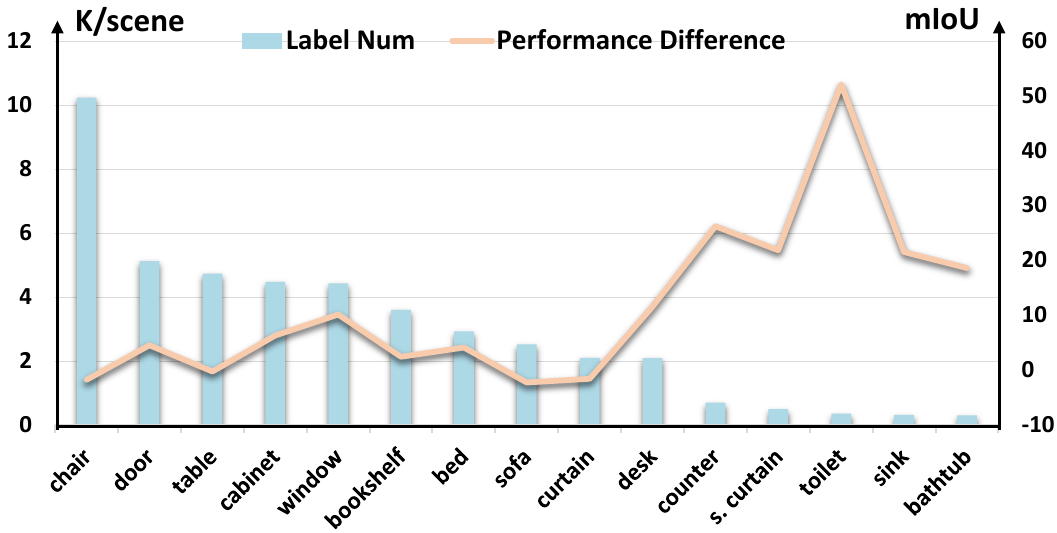}
\caption{\textbf{Difference of semantic segmentation performance for different categories.} The sorted histogram illustrates the average number of labeled points across various categories, while the line chart depicts the performance improvement deriving from switching from joint consistency regularization to instance consistency regularization. We only report the categories with respectable performance.}
\label{fig:loss-performance}
\vspace{-0.4cm}
\end{figure}
2) Though we do not employ semantic consistency loss upon unlabeled data in instance consistency regularization, they still show the best performance of semantic segmentation. The instance consistency regularization not only boosts instance segmentation but also helps discriminate semantic categories to obtain a well-balanced prediction. 

Meanwhile, as illustrated in Fig.~\ref{fig:loss-performance}, the semantic segmentation for weak categories with limited annotation scales in the right half is significantly improved. This proves that employing only instance pseudo labels can enable students to learn accurate semantic information.
\begin{figure*}[!t]
\centering
\includegraphics[width=1\linewidth]{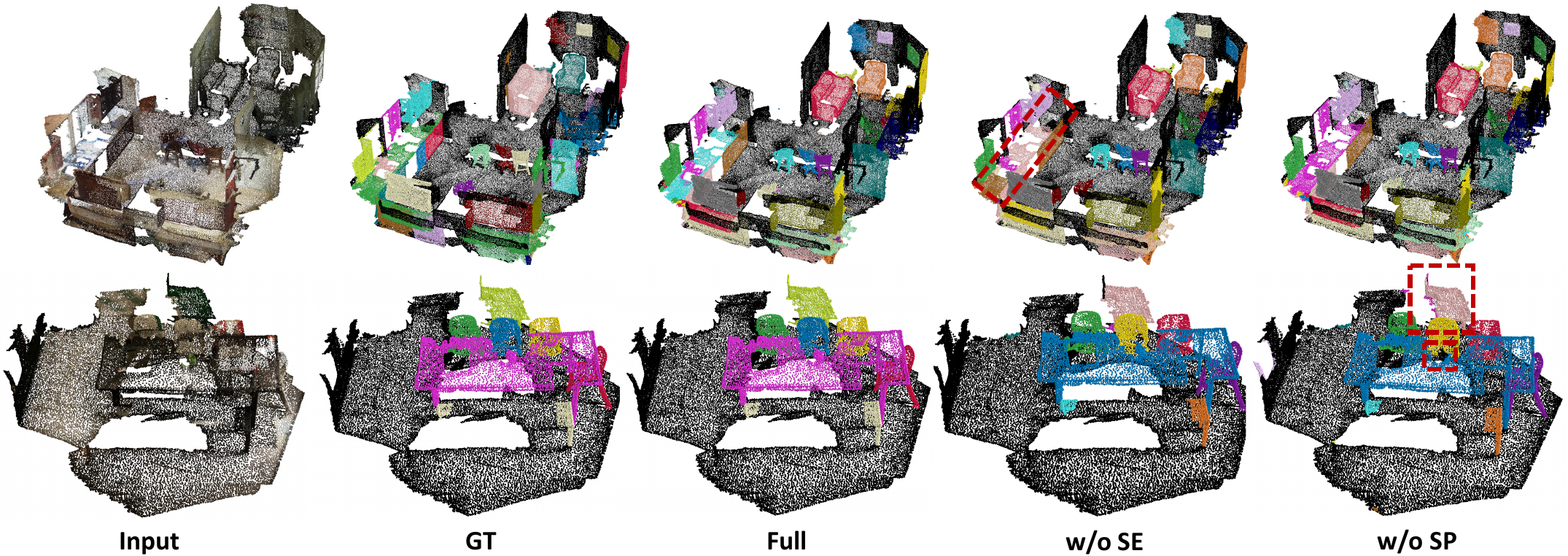}
\caption{\textbf{The qualitative results of dynamic mask generation module.} The bounding boxes highlight the key details. Best viewed in color.}
\label{fig:visualization-dmg}
\vspace{-0.2cm}
\end{figure*}

\subsubsection{Evaluation of segmentation models}\label{serial-parallel}
\noindent \textbf{Base segmentation models.}
We compare the state-of-the-art approaches of several instance segmentation paradigms in Table~\ref{table:superiority-dknet}. The group-based approaches serially rely on semantic prediction, which suffers from imbalanced category distribution, especially with limited annotation. The transformer-based and kernel-based approaches decode objects via encoded informative representation of instances. The transformer-based approaches focus on the design of the instance query encoder to build queries via learning from large-scale datasets, which leads to an inferior performance under a data-efficient setting. As shown in Table~\ref{table:superiority-dknet}, the performance of MAFT~\cite{lai2023mask} drops catastrophically when trained from scratch and with limited annotation. On the contrary, kernel-based approaches generate or assign kernels for instances depending on scenes. To build discriminative kernels, DKNet~\cite{dknet} thoroughly explores the data similarity to make full use of instance information, which demonstrates its superiority with $20\%$ labeled data in Table~\ref{table:superiority-dknet}. 

\noindent \textbf{Parallel segmentation framework.}
To evaluate the effects of the parallel segmentation framework, we build two serial models for comparison, where semantic and instance segmentation respectively rely on each other in two ways, termed ``sem2ins'' and ``ins2sem'', respectively. The model ``sem2ins'' implements a bottom-up segmentation model~\cite{hais} and adopts the mean teacher framework to realize semi-supervised instance segmentation. In the model ``ins2sem'', the point-wise semantic prediction is covered by broadcasting the instance-level categories predicted by an instance-wise classification MLP. Both serial approaches utilize joint instance and semantic consistency of unlabeled data.  
As we can observe in Table.~\ref{table:model-structure}, compared with the parallel segmentation model, both the above models show inferior performance, which demonstrates the superiority of parallel segmentation in semi-supervised learning. 

\begin{table}[!t]
\small
\begin{center}
\caption{Comparison of different base instance segmentation models. The previous state-of-the-art approaches are included.  ``*'' means that the models are trained from scratch. All the performances under the 20\% setting are obtained with only 20\% labeled data.}
\label{table:superiority-dknet}
\vspace{-0.4cm}
\begin{tabular*}{\linewidth}{@{}@{\extracolsep{\fill}}ll|cc|cc@{}}
\toprule
\multirow{2}*{Paradigm}&\multirow{2}*{Approach}&\multicolumn{2}{c|}{$100\%$}&\multicolumn{2}{c}{$20\%$}\\
~&~&mAP&$AP_{50}$&mAP&$AP_{50}$\\
\midrule
Group&HAIS~\cite{hais}&43.5 &64.1 &29.7 &48.9\\
Transformer  &MAFT~\cite{lai2023mask} &58.4&75.9 &-&- \\
Transformer  &MAFT* &50.3&68.1 &28.3&41.5 \\
Kernel  & DKNet~\cite{dknet} &50.8&66.7 &35.3&51.0 \\
\bottomrule
\end{tabular*}
\end{center}
\vspace{-0.4cm}
\end{table}
\begin{table}[!t]
\small
\begin{center}
\caption{Comparison of different segmentation architectures. ``JCR'' and ``ICR'' mean joint and instance consistency regularization, respectively.}
\label{table:model-structure}
\vspace{-0.4cm}
\begin{tabular*}{0.8\linewidth}{@{}@{\extracolsep{\fill}}c|ccc@{}}
\toprule
Architecture &mAP&$AP_{50}$&$AP_{25}$\\
\midrule
Serial (sem2ins) + JCR&$39.0$&$54.9$ &$66.5$  \\
Serial (ins2sem) + JCR&$34.7$&$52.5$ &$65.5$ \\
Parallel + JCR&$41.8$&$57.3$ &$67.5$ \\
Parallel + ICR &$\textbf{42.7}$&$\textbf{58.3}$ &$\textbf{68.2}$  \\
\bottomrule
\end{tabular*}
\end{center}
\vspace{-0.4cm}
\end{table}
\begin{table}[!t]
\small
\begin{center}
\caption{Evaluation of our centroid-focusing strategy. ``*'' means that the model is trained and evaluated with ground truth masks as an ideal representation.}
\label{table:ablation-centroid}
\vspace{-0.4cm}
\begin{tabular*}{0.8\linewidth}{@{}@{\extracolsep{\fill}}l|l|ccc@{}}
\toprule
Rate& Approach &mAP&$AP_{50}$&$AP_{25}$\\
\midrule
\multirow{2}*{$100\%$}&DKNet&50.8&66.7 &76.9  \\
~&DKNet*&51.5&67.0 &77.0 \\
\midrule
\multirow{3}*{$20\%$}&Random&39.5&56.2 &65.9 \\
~&IoU Match&40.9&57.1 &67.4\\
~&Ours&42.7&58.3&68.2\\
\bottomrule
\end{tabular*}
\end{center}
\vspace{-0.4cm}
\end{table}

\noindent \textbf{Effect of instance centroids.}
As depicted in Table~\ref{table:ablation-centroid}, we test an ideal representation (DKNet*) where features are collected from all the points within each ground-truth instance mask. It can be observed that the performance of DKNet is comparable with this ideal representation. This indicates that the centroid areas encapsulate sufficient discriminative instance contexts to build instance kernels. Besides, we build two models that deviate from focusing on centroid areas. The ``Random'' model randomly chooses candidates as instance centroids instead of those with the highest centroid scores. The ``IoU Match'' means that we replace centroid distances with IoUs between instance mask predictions and ground truth masks as matching costs to assign instance labels. Both the above two models show inferior performances, which reveals the potential of centroid areas in preventing noises.

\subsubsection{Effects of dynamic mask generation} To generate high-quality instance pseudo labels $\hat{M}_u$, we propose a dynamic mask generation (DMG) module including instance self-enhancement and superpoint refinement. 
We evaluate the effectiveness of different components of DMG, \emph{i.e}. instance self-enhancement (SE) and superpoint refinement (SP) in Table.~\ref{table:dynamic-generation}. 
Since InsTeacher3D, when not equipped with the DMG module (as shown in Row 1), already exhibits a convincing performance, incorporating DMG can significantly enhance its overall segmentation outcomes.
Firstly, the results presented in Row 2 demonstrate that the addition of the SP module leads to a noteworthy increase of $3.5$ in 
mAP. As mAP serves as a measure of the thoroughness and completeness of the segmented instance masks, this improvement confirms the effectiveness of superpoint refinement in enhancing the completeness of the initial mask predictions.
Secondly, the findings in Row 3 illustrate that the employment of the SE mechanism boosts the $AP_{50}$ metric to $57.0$. This substantial improvement attests to the ability of SE to effectively retain weakly expressed instances while eliminating instances that have been over-segmented. Ultimately, the synergistic effect of combining SE and SP is evidenced by the further enhancement in the quality of the generated pseudo instance labels. We utilize this collaborative strategy in InsTeacher3D to ensure more reliable semi-supervised learning, contributing to its stability and performance gains.
We also provide qualitative results to evaluate the effectiveness of different components as illustrated in Fig.~\ref{fig:visualization-dmg}.

\subsubsection{Ablation of Smoothing Factor in EMA}
After every step $\tau$, the weight of teacher model $\Phi^t$ is updated by the EMA weights of the student model $\Phi^t$, which can be formulated as Eqn.~\ref{eq:ema}.
This procedure can smooth the model noise and improve the performance of $\Phi^t$. To select a smoothing factor for better learning, we design an ablation experiment using different $\alpha$, the results of which are shown in Table.~\ref{table:smoothing_factor}. Observed that models trained with $\alpha$ larger than $0.999$ show a minor impact on performance, we set $\alpha$ as $0.999$ to maintain stable training.

\begin{table}[!t]
\small
\begin{center}
\caption{Comparison of different components in dynamic mask generation. SE and SP denote the instance self-enhancement module and superpoint refinement module, respectively.}
\label{table:dynamic-generation}
\vspace{-0.4cm}
\begin{tabular*}{0.7\linewidth}{@{}@{\extracolsep{\fill}}cc|ccc@{}}
\toprule
SE & SP &mAP&$AP_{50}$&$AP_{25}$\\
\midrule
\ding{55} &\ding{55} &$37.9$&$55.6$ &$66.4$ \\
\ding{55} & \checkmark &$41.4$&$56.4$ &$66.5$ \\
\checkmark & \ding{55}&$38.3$&$57.0$ &$67.7$ \\
\checkmark & \checkmark & $\textbf{42.7}$&$\textbf{58.3}$ &$\textbf{68.2}$  \\
\bottomrule
\end{tabular*}
\end{center}
\vspace{-0.4cm}
\end{table}
\begin{table}[!t]
\small
\begin{center}
\caption{Comparison of models trained with different smoothing factors.}
\label{table:smoothing_factor}
\vspace{-0.4cm}
\begin{tabular*}{0.7\linewidth}{@{}@{\extracolsep{\fill}}c|ccc@{}}
\toprule
$\alpha$ &mAP&$AP_{50}$&$AP_{25}$\\
\midrule
0.99&39.4&	55.7&	65.2\\
0.999&\textbf{42.7}&58.3 &\textbf{68.2} \\
0.9999&42.0& \textbf{58.5} & 68.1 \\
\bottomrule
\end{tabular*}
\end{center}
\vspace{-0.4cm}
\end{table}

\section{Conclusion}
In this paper, we argue that semantic pseudo labels are not sufficiently reliable and that leveraging only instance pseudo labels is preferable for semi-supervised 3D instance segmentation. We propose a novel self-training network, called InsTeacher3D, combining a parallel instance segmentation model DKNet with an instance consistency regularization framework. InTeacher3D effectively generates and leverages high-quality instance pseudo labels, facilitating semi-supervised 3D instance segmentation.
Our experimental results demonstrate that InsTeacher3D significantly outperforms previous state-of-the-art approaches, and a series of ablation studies provide evidence of the effectiveness of every component.


%




\ifCLASSOPTIONcaptionsoff
  \newpage
\fi



%
{\small
\bibliographystyle{IEEEtran}
\bibliography{egbib}
}
%








\end{document}